\documentclass[10pt]{article}
\usepackage[utf8]{inputenc}

\usepackage{authblk}

\usepackage{amsmath}

\newtheorem{heuristic}{Heuristic}

\usepackage[english]{babel}

\usepackage[numbers]{natbib}

\usepackage[hyphens]{url}
\usepackage{xcolor}
\usepackage{amsmath}
\usepackage{amssymb}
\usepackage{mathtools}
\usepackage{hyperref}
\hypersetup{
    colorlinks = true,
    urlcolor = black,
    linkcolor = black,
    citecolor = black
}

\usepackage{graphicx}
\usepackage{float}
\graphicspath{ {images/} }
\usepackage{caption}
\usepackage{subcaption}
\usepackage{tabulary}
\usepackage{longtable}

\usepackage[toc,page]{appendix}

\usepackage{dirtytalk}

\usepackage[ruled, noline]{algorithm2e}
\DontPrintSemicolon
\SetKwFor{ForEachUntil}{foreach}{do}{}
\SetKw{Until}{until}
\SetKw{With}{with}
\SetKw{Do}{do}
\SetKw{Goto}{go to}

\newcommand{\rank}{\rho}

\newcommand{\armexpreward}{\mu}
\newcommand{\reward}{r}
\newcommand{\configurations}{\mathcal{C}}
\newcommand{\banditepimodel}{\mathcal{E}}
\DeclareMathOperator*{\E}{\mathbb{E}}

\newcommand{\infections}{ARI}
\newcommand{\hospitalisations}{ARH}

\newcommand{\mRNA}{mRNA}
\newcommand{\adeno}{vector-based}

\newcommand{\scenarioZero}{\textit{Baseline}}
\newcommand{\scenarioOne}{\textit{Relaxed}}
\newcommand{\scenarioTwo}{\textit{Tertiary Education}}
\newcommand{\scenarioThree}{\textit{Secondary Schools}}
\newcommand{\scenarioFour}{\textit{Relaxed Community}}
\newcommand{\scenarioFive}{\textit{Relaxed Workplace}}

\newtheorem{definition}{Definition}

\newcommand{\supplementary}[1]{Appendix~#1}
\newcommand{\tc}[1]{\textcolor{black}{#1}}

\begin{document}

\title{Evaluating COVID-19 vaccine allocation policies using Bayesian m-top exploration}
\author[1]{Alexandra Cimpean}
\author[1]{Timothy Verstraeten} 
\author[3,4]{Lander Willem} 
\author[2,3]{Niel Hens} 
\author[1]{Ann Now\'{e}} 
\author[1,2]{Pieter Libin}
\affil[1]{\small Artificial Intelligence Lab, Department of Computer Science, Vrije Universiteit Brussel, Brussels, Belgium}
\affil[2]{Data Science Institute, Interuniversity Institute of Biostatistics and statistical Bioinformatics, UHasselt, Hasselt, Belgium}
\affil[3]{Centre for Health Economics Research and Modelling Infectious Diseases, Vaccine \& Infectious Disease Institute, University of Antwerp, Antwerp, Belgium
}
 \affil[4]{Department of Family Medicine and Population Health (FAMPOP), University of Antwerp, Antwerp, Belgium}

\date{}

\maketitle

\begin{abstract}
\noindent
Individual-based epidemiological models support the study of fine-grained preventive measures, such as tailored vaccine allocation policies, \textit{in silico}. As individual-based models are computationally intensive, it is pivotal to identify optimal strategies within a reasonable computational budget. Moreover, due to the high societal impact associated with the implementation of preventive strategies, uncertainty regarding decisions should be communicated to policy makers, which is naturally embedded in a Bayesian approach.

We present a novel technique for evaluating vaccine allocation strategies using a multi-armed bandit framework in combination with a Bayesian anytime $m$-top exploration algorithm. $m$-top exploration allows the algorithm to learn $m$ policies for which it expects the highest utility, enabling experts to inspect this small set of alternative strategies, along with their quantified uncertainty. The anytime component provides policy advisors with flexibility regarding the computation time and the desired confidence, which is important as it is difficult to make this trade-off beforehand.

We consider the Belgian COVID-19 epidemic using the individual-based model STRIDE, where we learn a set of vaccination policies that minimize the number of infections and hospitalisations.
Through experiments we show that our method can efficiently identify the $m$-top policies, which is validated in a scenario where the ground truth is available. Finally, we explore how vaccination policies can best be organised under different contact reduction schemes and we investigate the impact of vaccine uptake proportions (i.e., the proportion of individuals that will comply with the strategy and take the vaccine).
Through these experiments, we show that the top policies follow a clear trend regarding the prioritised age groups and assigned vaccine type, which provides insights for future vaccination campaigns. Furthermore, our experiments suggest that the uptake proportion has only a limited influence on overall policy optimality.

\end{abstract}

\section{Introduction}
Epidemiological models (e.g., compartment models and individual-based models) are essential to study the effects of preventive measures \textit{in silico} \cite{Basta2009,Germann2006}.
While individual-based disease transmission models are typically associated with greater model complexity and computational cost than compartment models, they allow for a more realistic evaluation of preventive strategies \cite{Eubank2006}, provided they are well informed \cite{willem2017lessons}.
To capitalize on these advantages and to make it feasible to employ large-scale individual-based models, it is essential to use the computational resources as efficiently as possible.

In the literature, a set of possible preventive strategies is typically evaluated by simulating each of the strategies an equal number of times \cite{Fumanelli2016,ferguson2005strategies,Chao2012}.
However, this approach is computationally inefficient to identify the optimal preventive strategies, as a large proportion of computational resources will be used to evaluate sub-optimal strategies.
Furthermore, a consensus on the required number of model evaluations per strategy is currently lacking \cite{Willem2014} and it was shown that this number depends on the \emph{hardness} of the evaluation problem \citep{Libin2018}.
As running an individual-based model is computationally intensive (i.e., minutes to hours, depending on the complexity of the model), minimizing the number of required model evaluations reduces the total time required to evaluate a given set of preventive strategies. This renders the use of individual-based models attainable in studies that would otherwise not be computationally feasible. Additionally, reducing the number of model evaluations will free up computational resources in studies that already use individual-based models, capacitating researchers to explore a broader set of model scenarios. This is important, as considering a wider range of scenarios increases the confidence about the overall utility of preventive strategies \cite{wu2006reducing}.

In this work, we present a novel technique to evaluate preventive strategies using a multi-armed bandit framework in combination with anytime $m$-top exploration algorithms. 
By formulating the decision problem as an $m$-top exploration bandit problem, a learning agent can select the $m$ policies for which it expects the highest utility, enabling experts to inspect this small set of alternatives.
The anytime component provides the policy advisors with flexibility regarding the time at which a decision is made. 
This is especially important when computationally intensive models are used as for such models it is difficult to make a trade-off between the available budget and desired confidence.
We focus on a Bayesian learning approach, to quantify the uncertainty of the decision making.

Using this innovative framework, we study a vaccine allocation problem, where we investigate how the weekly supply of COVID-19 vaccines in Belgium could have been optimally allocated to the different age groups in the population. As vaccines are administered gradually, certain contact reductions remained in place during the vaccination campaign to curb the disease burden. Whether the design of social contact restrictions affects the optimal vaccine allocation, is part of our experimental exploration. Moreover, we investigate the impact of the vaccine uptake proportion (i.e., the proportion of individuals that will comply with the strategy and take the vaccine) on the design of vaccine allocation strategies \cite{Pertwee2022}. In this regard, we study the impact of household clustering of unvaccinated individuals \cite{Kuylen2020}. 
To support detailed contact reduction schemes and investigate vaccine uptake at the household level, the use of a fine-grained individual-based model is warranted \cite{Kuylen2020,Libin2021b,Chan2024}.

\section{Related Work}
Epidemic control has been explored in a reinforcement learning setting, both from a stateful and a multi-armed bandit perspective. 

From a stateful reinforcement learning perspective, the concept of learning dynamic policies by formulating the decision problem as a Markov decision process (MDP) was first introduced by Yaesoubi and Cohen \citep{Yaesoubi2011}. To investigate dynamic tuberculosis case-finding policies in HIV/tuberculosis co-epidemics, a policy iteration algorithm was used to solve the MDP \citep{Yaesoubi2013}. This technique was later extended to include cost-effectiveness in the analysis and applied to mitigation policies (that is, school closures and vaccines) in the context of pandemic influenza in a simplified epidemiological model \citep{Yaesoubi2016}. More recently, Libin et al. used deep reinforcement learning to learn mitigation strategies in the context of pandemic influenza \citep{Libin2021}. Reymond et al. explored COVID-19 mitigation policies from a multi-objective reinforcement learning perspective, where complex mitigation policies with possibly conflicting objectives are balanced to learn the best trade-offs \citep{Reymond2022}.

From a multi-armed bandit perspective, we distinguish efforts that investigate a cumulative regret and a best-arm identification setting. On the one hand, in the cumulative regret setting, we identified work focusing on various preventive strategies in the context of COVID-19 \citep{Awasthi2022,Grushka-Cohen2020,Bastani2021}. We note that these studies do not consider individual-based models. On the other hand, best-arm identification algorithms have been used to evaluate preventive strategies in individual-based models \citep{Libin2018}, which we consider the work most closely related to our study. In that work, Bayesian fixed-budget best-arm identification algorithms\footnote{A broad overview on the state of the art with respect to (Bayesian) best-arm identification algorithms is provided by Kaufmann et al. and Hoffman et al. \cite{kaufmann2016complexity,hoffman2014correlation}.} such as Top-two Thompson sampling have been used before to evaluate preventive strategies in the context of pandemic influenza \citep{Libin2018}. 

However, the use of fixed-budget best-arm identification has some important limitations. First, simply returning the single best prevention strategy can be an obstacle for public health scientists, as this implies that public health scientists can only offer a take-it-or-leave-it option to government officials, rather than a set of options that can be evaluated within the political and legal framework of the government.  Additionally, from a health economics perspective, a set of optimal policies can be used to negotiate a fair cost with the producers of pharmaceutical supplies. Second, Libin et al. assume a fixed computational budget, that needs to be specified a priori \citep{Libin2018}. We argue that deciding the budget upfront can be challenging, which is especially the case when computationally expensive models are used, for which it is difficult to make a trade-off between the available budget and desired confidence. As such, we assert that an anytime bandit setting can overcome these limitations, as an initial budget can still be provided, but the budget can be extended when necessary. To address these limitation, in this work, we study the anytime $m$-top exploration problem, introduced by Jun et al. \cite{jun2016}. Jun et al. introduce the frequentist algorithm AT-LUCB \cite{jun2016}. We note that as a UCB-variant, AT-LUCB is not equipped to incorporate prior knowledge with respect to the reward distribution. As Libin et al. have shown that incorporating such knowledge can greatly improve the learning performance \citep{Libin2018}, we study a Thompson sampling algorithm to solve the anytime $m$-top exploration problem: Boundary Focused Thompson sampling \cite{Libin2019}.

\section{Methods}
\subsection{Epidemic bandits}
\label{sec:methods}
We formulate the evaluation of preventive strategies as a multi-armed bandit problem \citep{Libin2018}, with the aim of identifying the $m$-top arms using anytime decision making algorithms \citep{Libin2019}. The presented method is generic, capable of dealing with different epidemic model types, that consider distinct pathogens, contact networks and preventive strategies. This method will be evaluated in the context of COVID-19 in the next section.

First, we formally define the multi-armed bandit.
\begin{definition}[Multi-armed bandit]
\label{def:bandit}
\upshape A \emph{multi-armed bandit} involves $K$ arms that can be pulled \cite{audibert2010best}, where each arm $a_k$ has a \emph{reward distribution}. When an arm $a_k$ is pulled, it returns a reward $r_k$ sampled from $a_k$'s reward distribution. For each arm $a_k$ we have the expected reward $\armexpreward_k = \E\left[\reward_k\right]$.
\end{definition}

A common use of the multi-armed bandit is to pull a sequence of arms such that the best arm is identified.
However, in this work, our aim is to solve the $m$-top exploration problem ($m < K$), where the objective is to identify the $m$ best arms, with respect to the expected reward $\armexpreward_k$ of the arms \citep{Bechhofer1958}.
Formally, we have $\armexpreward_1 \geq \ldots \geq \armexpreward_m \geq \armexpreward_{m+1} \geq \ldots \geq \armexpreward_K$, and the objective is to identify the set $\{\armexpreward_1, \ldots,  \armexpreward_m \}$.
This is a \textit{pure exploration} problem where the focus is on gaining knowledge about which $m$ arms are ranked the highest.

Next, we provide a formal definition of the epidemic model we consider \citep{Libin2018}.
\begin{definition}[Stochastic epidemiological model]
\label{def:stoch_epi_model}
\upshape  A \emph{stochastic epidemiological model} $\banditepimodel$ is defined in terms of a model configuration $c \in \configurations$ and can be used to evaluate a preventive strategy $p$.
  The result of a model evaluation is referred to as the \emph{model outcome}.
  Evaluating the model $\banditepimodel$ thus results in a sample of the model's \emph{outcome distribution}:
  \begin{equation}
   \text{outcome} \sim \banditepimodel(c, p)
  \end{equation}
\end{definition}
The model outcome can be any statistic relevant to the decision maker, such as prevalence, proportion of symptomatic individuals, proportion of hospitalised individuals, mortality or societal cost.
Note that a model configuration $c \in \configurations$ describes the complete model environment, i.e., both aspects inherent to the model and options that the modeller can provide (e.g., population statistics, vaccine properties).

Our objective is to find the set of $m$-top preventive strategies (i.e., the strategies that minimize the expected outcome) from a set of alternative strategies
\begin{equation}
\{p_1,...,p_K\},
\end{equation} 
for a particular configuration \begin{equation}
c_0 \in \configurations,
\end{equation}
where $c_0$ corresponds to the context of the studied epidemic.
To this end, we consider a multi-armed bandit with preventive strategies $\{p_1,...,p_{K}\}$ represented by arms $\{a_1,...,a_{K}\}$. Pulling arm $a_k$ corresponds to evaluating the corresponding preventive strategy $p_k$, by running a simulation in the epidemiological model $\banditepimodel(c_0, p_k)$. The bandit thus has preventive strategies as arms with reward distributions corresponding to the outcome distribution of an epidemiological model $\banditepimodel(c_0, p_k)$.
While the parameters of the outcome distribution (i.e., the parameters of the epidemiological model) are known, it is intractable to determine the top strategies analytically. Hence, we must learn about the outcome distribution via interaction with the epidemiological model.

\subsection{$m$-top exploration}
\label{sec:best_arm_id}

Our objective is to identify the $m$-top preventive strategies for a particular configuration of an epidemiological model.
We consider two anytime $m$-top algorithms: AnyTime Lower and Upper Confidence Bound (AT-LUCB) and Boundary Focused Thompson Sampling (BFTS).

\subsubsection*{AnyTime Lower and Upper Confidence Bound algorithm}
The AT-LUCB algorithm invokes the fixed-confidence LUCB algorithm \cite{jun2016,kalyanakrishnan2012pac}.
At each time step $t$, AT-LUCB (Algorithm \ref{algorithm:AT-LUCB}) returns the empirical $m$-top arms $J^{(t)}$. Given $K$ number of arms, $\hat{\mu}_a^{(t)}$ is the empirical mean for arm $a$ at time step $t$. The amount of times arm $a$ was pulled until time $t$ is denoted by $n_a^{(t)}$. Given the LUCB stage index $s$, the confidence parameter at stage $s$ is determined by a decaying failure parameter $\delta_s=\delta_1\alpha^{(s-1)}$. The stage to which time $t$ belongs is defined as $S^{(t)}$.
\\\noindent
\begin{minipage}[H]{\linewidth}
\vspace{\baselineskip}
\begin{algorithm}[H]
\caption{AT-LUCB}\label{algorithm:AT-LUCB}
\KwIn{$\delta_1 \leq [1/200, K]$, $\alpha \in [1/50,1)$, $\epsilon \geq 0$}
\;
$S^{(0)} \leftarrow 1$ \;
$\delta_s \leftarrow \delta_1\alpha^{(s-1)},\quad \forall s \geq 1$ \;
\For{$t=1,\dots,+\infty$}{
    \eIf{$\mathrm{Term}^{(t)}(\delta_{S^{(t-1)}}, \epsilon)$ }{
        $S^{(t)} \leftarrow \max \{ s' \geq S^{(t-1)}+1 : \neg \mathrm{Term}^{(t)}(\delta_{s'}, \epsilon) \}$ \;
        $J^{(t)} \leftarrow $ \{the empirical $m$-top arms\} \;
    }{
        $S^{(t)} \leftarrow S^{(t-1)}$ \;
        $J^{(t)} \leftarrow J^{(t-1)}$ (or empirical $m$-top arms if $S^{(t)}=1$) \;
    }
    Pull $h_*^{(t)}(\delta_{S^{(t)}})$ and $l_*^{(t)}(\delta_{S^{(t)}})$ as in Equation \ref{eq:atlucb_arms}\;
    Recommend $J^{(t)}$ \;
}
\end{algorithm}
\vspace{\baselineskip}
\end{minipage}
The exploration strategy of AT-LUCB relies on the upper confidence bound $U_a^{(t)}$ and lower confidence bound $L_a^{(t)}$:
\begin{equation}\label{eq:bound}
\begin{split}
&U_a^{(t)}(\delta_s) = \hat{\mu}_a^{(t)} + \beta(n_a^{(t)},t,\delta_s)\\
&L_a^{(t)}(\delta_s) = \hat{\mu}_a^{(t)} - \beta(n_a^{(t)},t,\delta_s),
\end{split}
\end{equation}
with, 
\begin{equation}
\beta(n_a^{(t)},t,\delta_s)=\sqrt{\frac{1}{2n_a^{(t)}}\ln\left(\frac{5}{4}\frac{K \cdot t^4}{\delta_s}\right)}.
\end{equation}
Each time step $t$, the algorithm pulls arms 
\begin{equation}
\label{eq:atlucb_arms}
\begin{split}
&h_*^{(t)}(\delta_{S^{(t)}}) = \arg\min_{a \in \mathrm{High}^{(t)}}L^{(t)}_a(\delta)\\
&l_*^{(t)}(\delta_{S^{(t)}}) = \arg\max_{a \in \mathrm{High}^{(t)}}U^{(t)}_a(\delta),
\end{split}
\end{equation}
with $\mathrm{High}^{(t)}$ the $m$-top arms at time $t - 1$. 
When the terminating condition $\mathrm{Term}^{(t)}(\delta, \epsilon) = \{ U^{(t)}_{l_*^{(t)}(\delta)}(\delta) - L^{(t)}_{h_*^{(t)}(\delta)}(\delta) < \epsilon \}$ is met, the algorithm moves to the next stage.

\subsubsection*{Boundary Focused Thompson Sampling}
\label{section:bayesian_sampling}

While confidence bound algorithms such as AT-LUCB permit specifying tight theoretical bounds, algorithms based on Thompson sampling typically perform better in practice \cite{chapelle2011empirical}.
Thompson sampling uses samples of the bandit's posteriors to decide which arm to pull next.

By using a Bayesian $m$-top identification algorithm, prior knowledge about the outcome distributions can be taken into account when defining an appropriate prior and posterior on the arms' reward distributions. This prior knowledge can increase the sample efficiency while the resulting posteriors provide valuable information about the decision uncertainty to guide policy makers.

For a multi-armed bandit, our prior belief over the arms' means is given by a prior distribution $\pi(.)$. Given an observed history $\mathcal{H}^{(t-1)}$ of rewards $r$ from pulling arms $a$ until timestep $t - 1$, where
\begin{equation}
    \nonumber \mathcal{H}^{(t-1)} = \left\{ a^{(i)}, r^{(i)}\right\}_{i=1}^{(t-1)},
\end{equation}
the posterior over the means of the bandit is defined as:
\begin{equation}
    \nonumber \pi \left( \cdot\ |\ \mathcal{H}^{(t-1)}  \right),
\end{equation}
where $\pi(\cdot)$ is conditioned on the observed history.

At each timestep $t$, Thompson sampling samples an estimate $\Tilde{\mu}^{(t)}_k$ of the mean $\mu_k$ from each posterior $k$ and ranks these samples to select the arm with the highest sampled mean.
By sampling from the posterior, Thompson sampling uses the uncertainty of the mean to balance exploration and exploitation.
As it is playing an arm multiple times, the posterior's uncertainty decreases and Thompson sampling will gear towards the highest ranking arms.
As Thompson sampling is able to use prior knowledge, sampling efficiency can be greatly improved.
In the context of epidemic decision making, we can derive such knowledge using epidemiological modelling theory, which we will do for the experimental scenario considered in Section~\ref{sec:experiments}.

\begin{figure}[h]
    \centering
    \includegraphics[width=0.5\linewidth, keepaspectratio]{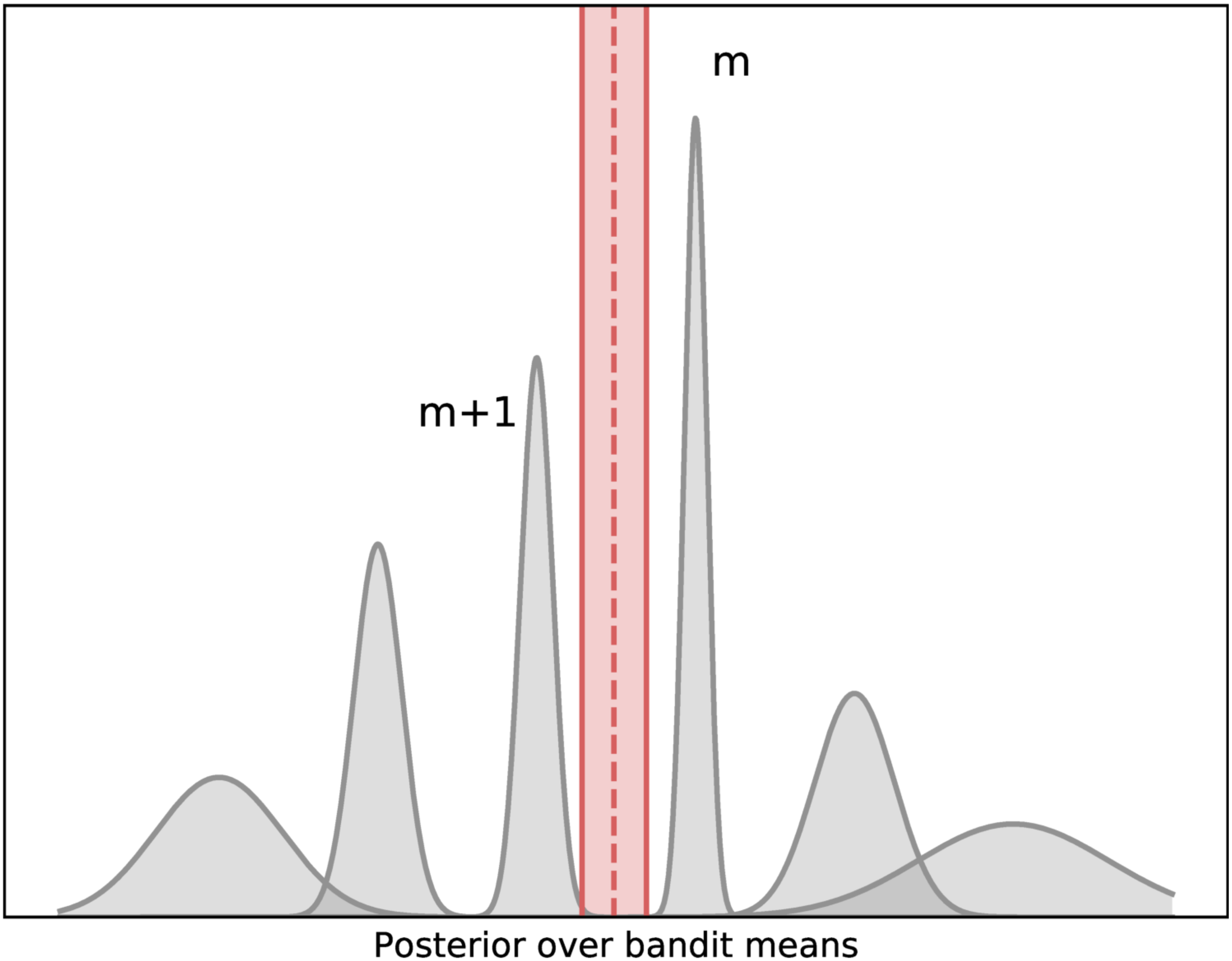}
    \caption{Posteriors for an artificial bandit ($K = 6$, $m = 3$) (gray) and BFTS’ decision boundary (red) with confidence bounds to demonstrate its uncertainty}
    \label{figure:BFTS_confidence_bound}
\end{figure}

Boundary Focused Thompson Sampling (BFTS) \citep{Libin2019} implements a Thompson sampling variant for $m$-top exploration. It uses the posterior samples as an estimate for the arms' means, which are ranked as in Thompson sampling.
BFTS strives to recommend the $m$-top best arms at any given time.
To denote the rank of the $\rho$-ordered arm, we define this operator:
\begin{equation}\label{eq:ts_ranking_operator}
    \psi_\rank(\mathbf{\Tilde{\mu}}^{(t)}).
\end{equation}
BFTS (Algorithm \ref{algorithm:BFTS}) focuses on both sides of the decision boundary for the $m$-top arms, in order to decrease the uncertainty about arms $a_m^{(t)}$ and $a_{m+1}^{(t)}$ with rankings $\psi_{m}(\mathbf{\Tilde{\mu}}^{(t)})$ and $\psi_{m+1}(\mathbf{\Tilde{\mu}}^{(t)})$, respectively.
Therefore, the arms ranked $\psi_{m}(\mathbf{\Tilde{\mu}}^{(t)})$ and $\psi_{m+1}(\mathbf{\Tilde{\mu}}^{(t)})$ are played with equal probability using a Bernoulli experiment. 

A key insight regarding BFTS is that its exploration is guided by sampling from the posterior distribution, balancing between $\psi_{m}(\mathbf{\Tilde{\mu}}^{(t)})$ and $\psi_{m+1}(\mathbf{\Tilde{\mu}}^{(t)})$, which represent our belief about the decision boundary at time $t$. Since the posterior captures the uncertainty inherent in the bandit problem, sampling from the $m^{\mathrm{th}}$ or ${m+1}^{\mathrm{th}}$ ordered arm initially promotes exploration across all arms when an uninformative prior is used. Over time, as the uncertainty for the outermost arms decreases, BFTS shifts its focus toward the arms closer to the decision boundary. Figure \ref{figure:BFTS_confidence_bound} illustrates this progression in a simple bandit scenario ($K=6$ and $m=3$) using Gaussian posteriors.
In \supplementary{B} we provide a Bayesian analysis of BFTS. While this analysis does not result in a bound on the simple regret, it does provide additional insight in BFTS’ exploration strategy and confirms that this strategy is well-grounded.
\\\noindent
\begin{minipage}[H]{\linewidth}
\vspace{\baselineskip}
\begin{algorithm}[H]
\caption{Boundary Focused Thompson Sampling}\label{algorithm:BFTS}
\KwIn{$\pi(\cdot)$, $\mathcal{H}^{(0)} = \emptyset$}
\;
\For{$t=1,\dots,+\infty$}{
    $\mathbf{\Tilde{\mu}}^{(t)} \sim \pi \left( \cdot\ |\ \mathcal{H}^{(t-1)} \right)$\;
    $b \sim \mathcal{B}er(0.5)$ \;
    $a^{(t)} = \psi_{m+b}(\mathbf{\Tilde{\mu}}^{(t)})$ \;
    $r^{(t)} \leftarrow \mathrm{Pull\ arm\ } a^{(t)}$ \;
    $\mathcal{H}^{(t)} \leftarrow \mathcal{H}^{(t-1)} \cup \left\{ a^{(t)}, r^{(t)} \right\}$ \;
    Recommend top arms based on $\pi \left( \cdot\ |\ \mathcal{H}^{(t-1)}  \right)$ \;
}
\end{algorithm}
\vspace{\baselineskip}
\end{minipage}

\section{Vaccine policy evaluation}
\label{sec:experiments}
SARS-CoV-2 has highlighted the importance of pandemic mitigation strategies \citep{Mafalda2022}.
This virus manifests in distinct clinical outcomes, ranging from asymptomatic infection to COVID-19 disease, which may induce mild to severe symptoms \citep{Wang2020, Mafalda2022}. Severe COVID-19 cases require hospitalisation and might result in a fatal outcome \citep{WHO,Mafalda2022}.
Up to November 2024, over 776 million confirmed cases and 7 million deaths were reported \cite{WHO2024}.
Since the end of 2019, new variants of SARS-CoV-2 have emerged. The first major mutation, D614G, induced an increased transmissibility and infectiousness, making it the dominant strain of the virus globally \citep{Zhou2021}. Subsequently, a series of Variants of Concern emerged which further increased transmissibility and/or disease severity \citep{CDC_VoC}.

To avoid the overflow of hospitals and to reduce mortality, measures to reduce the number of infections were taken.
In the first phase of the pandemic, such interventions were limited to imposing contact reductions \citep{Abrams2021}. In a later phase of the pandemic, i.e., begin 2021, vaccines became available in many countries \citep{Mafalda2022, Willem2021, Bettini2021}.

In this work, we focus on learning optimal policies to allocate vaccines to a large population when vaccines become available in limited batches, due to the gradual production of vaccines.
As vaccines are administered gradually, certain contact reductions need to be kept in place during the vaccination campaign to maintain the disease burden. However, the design of these social contact restrictions, including their focus and intensity, can vary while still maintaining comparable levels of disease burden. Therefore, we evaluate vaccine allocation strategies under different contact reduction scenarios in our experiments.
We investigate how to organize COVID-19 vaccine allocation policies targeted at the minimisation of two distinct criteria: infections and hospitalisations.
We explore different determinants regarding vaccine allocation policies, including the targeted age group and the vaccine type (i.e., mRNA and vector-based), which results in a large number of preventive policies that is to be evaluated.

We consider the Belgian COVID-19 epidemic in early 2021, where vaccine supplies started to be delivered on a weekly basis, with a changing supply rate as vaccine production increased over time.
We take into account the two types of vaccines that were available in Belgium during this phase, i.e., mRNA \citep{Polack2020, ZhangNa2020} and vector-based vaccines \citep{Zhu2020}. We investigate how a weekly supply of vaccines is best allocated among all age groups of the population.
We consider different social distancing schemes, under distinct vaccine uptake proportions (i.e., the proportion of individuals that will comply with the strategy and take the vaccine), to explore the effect of such policies on the vaccination campaign. 
Specifically, we study the impact of household clustering in vaccine uptake \cite{Kuylen2020}. Household clustering is important in this regard, as recent work shows that households constitute a reasonable proxy for predictors associated with vaccine hesitancy \cite{Chaudhuri2022}. Moreover, parents who have a negative attitude towards vaccination might be reluctant to vaccinate their children \cite{Kuylen2020}. 

Children were excluded from the initial COVID-19 vaccination campaigns in 2021 for regulatory reasons. However, they are considered vaccine eligible in our study to enable a population-wide assessment to shape future vaccine allocation strategies, consistent with earlier research \cite{Medlock2009,angeli2025insights}.

To support detailed contact reduction schemes and investigate vaccine
uptake at the household level, the use of a fine-grained individual-based model
is warranted  \cite{Kuylen2017, Kuylen2020, Chan2024}. To this end, we use the STRIDE
individual-based simulator \cite{Kuylen2017}, to explicitly model 11 million Belgians \cite{Willem2021}, that can engage in social contacts at home, in workplaces, in schools or in the general community.
To enable a Bayesian learning approach, we will introduce priors for these scenarios using insights from epidemic modelling theory in Section~\ref{section:outcome_posteriors}.

\subsection{STRIDE model and configuration}

In our experiments, we start the simulation period on January 1st 2021, when the first COVID-19 vaccines became available and the circulating variant in Belgium was the Alpha VoC.
We use the individual-based model STRIDE to simulate the entire Belgian population of 11 million individuals. A single simulation considers 4 calendar months, from January 1st 2021 until May 1st 2021, and includes school holidays (January 1st to January 3rd, February 15th to February 21st and April 5th to April 18th). Any chosen vaccination strategy is fixed throughout the simulation, resulting in an aggregate reward at the end of the simulation.
Depending on the social contact scenario, a distinct regimen of social contact reductions is imposed on the population. Imposing a higher contact reduction means individuals can participate in fewer person-to-person contacts, thereby reducing their likelihood to acquire infection.
We consider different social contact scenarios, as specified in Table \ref{table:configs}, to explore whether the vaccination strategy is affected by imposed contact reductions. 
The model explicitly accounts for contact tracing that was in place and additional details on this can be found in \supplementary{C}.

We use the STRIDE model configuration as calibrated on the Belgian COVID-19 epidemic in earlier work \cite{Willem2021}, where the first wave of the COVID-19 pandemic and the exit strategies were studied. Because we simulate the progress of the pandemic starting from January 1st 2021 and not from the start of the pandemic, the population is initialised with the proportion of immunity that was estimated at that moment in Belgium.
This proportion of immunity was estimated using the stochastic compartment model by Willem et al. \citep{Willem2024impact}.
We consider the Alpha VoC variant of SARS-CoV-2, which is 50\% more infectious compared to previously circulating variants \citep{tao_biological_2021}.


\begin{table}[ht]
\centering
  \begin{center}
    \begin{tabular}{ || p{1.82cm} || p{1.1cm} | p{1.4cm} | p{1.1cm} | p{1.5cm} | p{1.65cm} || }    \hline\hline
    \textit{Scenario Name} & Primary School & Secondary School & Tertiary School & Workplace & Community \\\hline\hline
    \scenarioZero & 0\% & 50\% & 100\% & 70\% & 70\% \\\hline
    \scenarioOne & 0\% & 50\% & 100\% & 50\% & 50\% \\\hline
    \scenarioTwo & 0\% & 50\% & 70\% & 70\% & 70\% \\\hline
    \scenarioThree & 0\% & 0\% & 100\% & 70\% & 70\% \\\hline
    \scenarioFour & 0\% & 50\% & 100\% & 70\% & 50\% \\\hline
    \scenarioFive & 0\% & 50\% & 100\% & 50\% & 70\% \\\hline
    \hline
    \end{tabular}
  \end{center}
  \caption{Social contact reduction schemes for the epidemic COVID-19 scenarios for Belgium. 0\% implies there is no reduction in contacts and 100\% means imposing full contact reduction.}
  \label{table:configs}
\end{table}

\subsection{Vaccine allocation}
\label{section:vaccine_allocation}

\begin{figure}[h!]\centering
  \begin{subfigure}[h!]{0.495\linewidth}\centering
    \includegraphics[width=\linewidth]{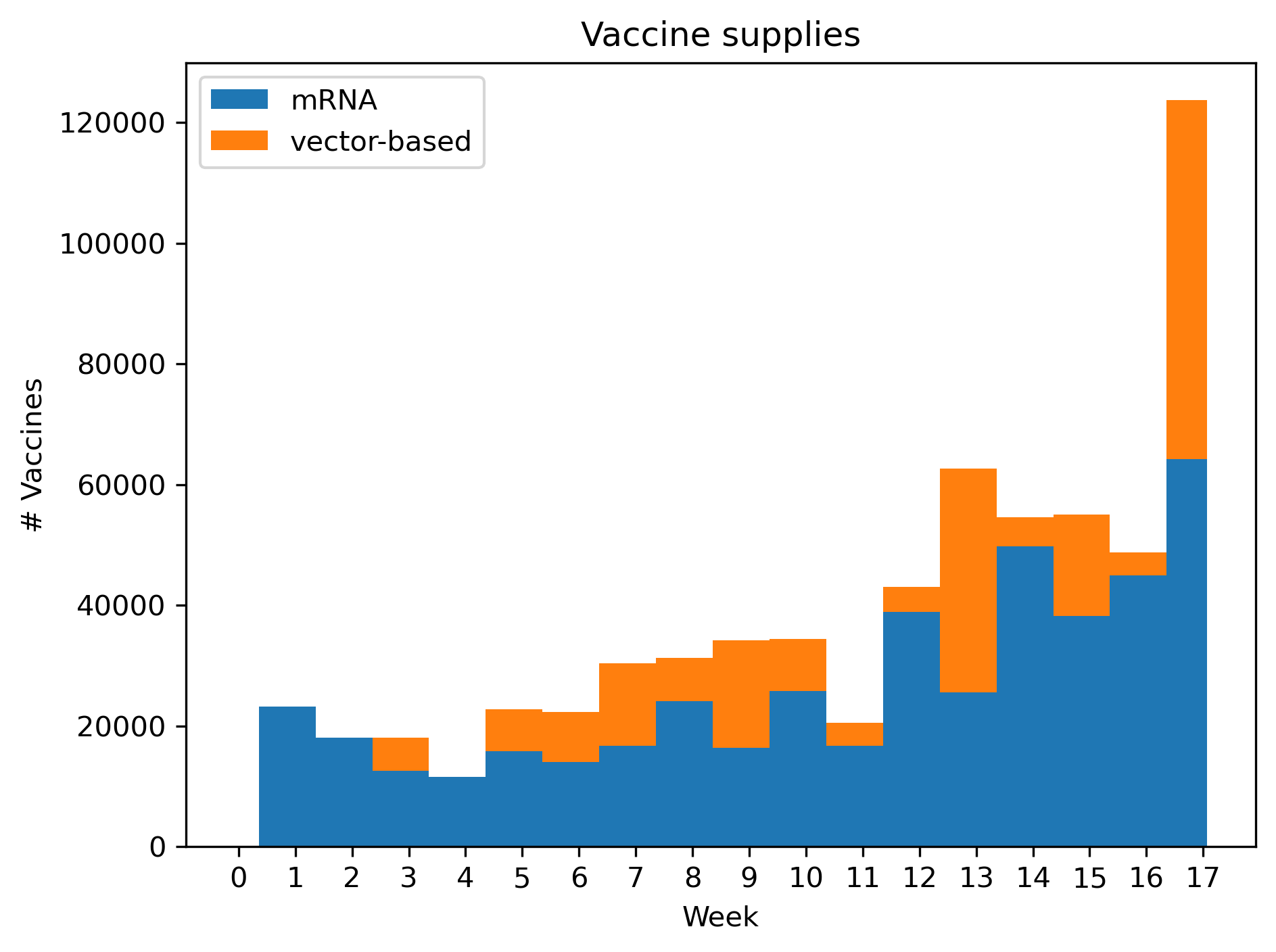}
    \caption{}
    \label{figure:vaccine_supplies}
  \end{subfigure}
    \begin{subfigure}[h!]{0.495\linewidth}\centering
    \includegraphics[width=\linewidth]{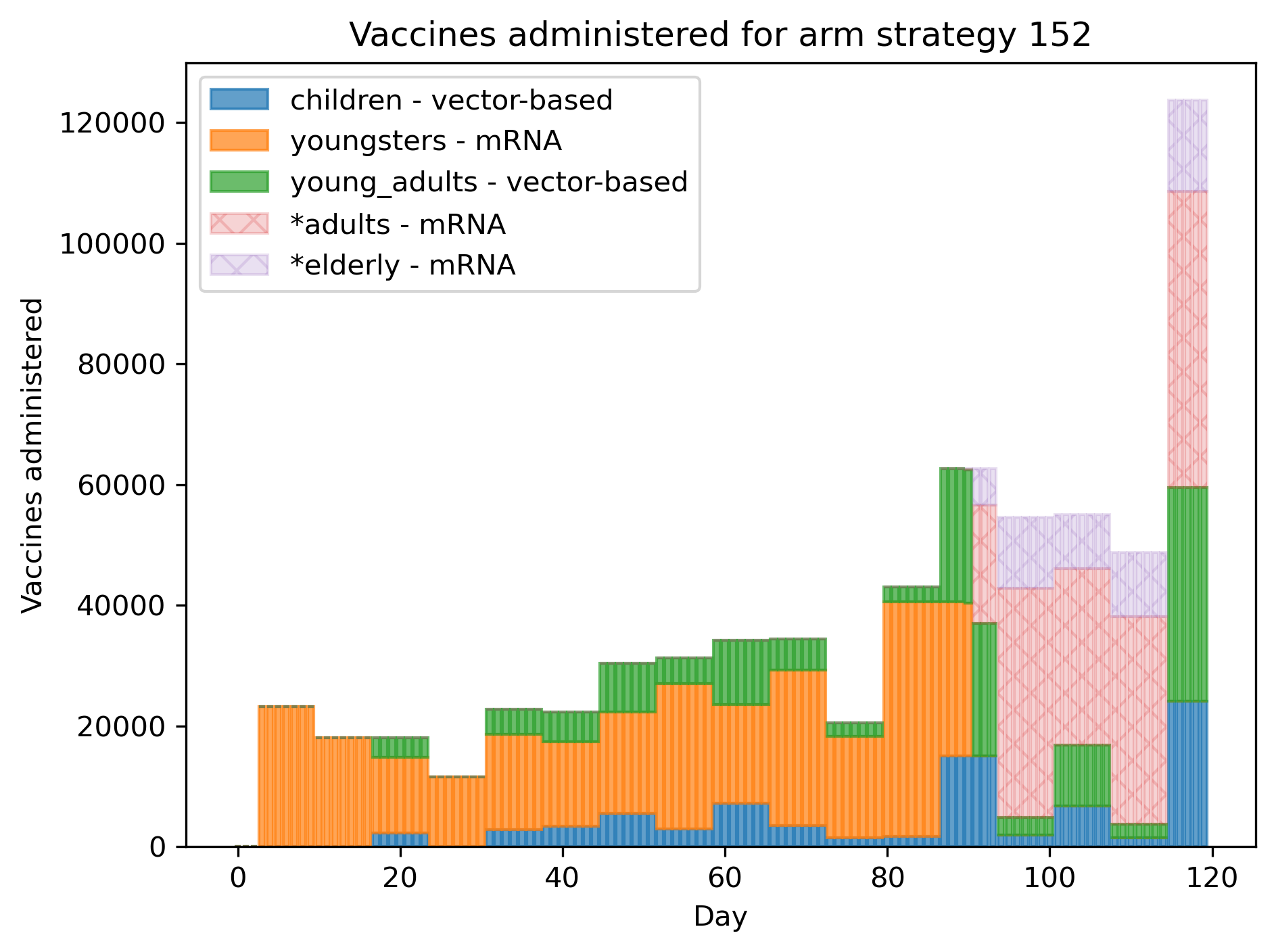}
    \caption{}
    \label{figure:vaccine_distribution}
  \end{subfigure}
  \caption{(a) Stacked bar chart of the reported vaccine supply from January 1st 2021 \citep{Vaesen2022}. (b) Stacked bar chart for one example of an uptake strategy where the entire populations accepts vaccines, starting with vaccinating children and young adults with vector-based vaccines and youngsters with mRNA vaccines. When the youngsters are fully vaccinated, remaining and newly arrived mRNA vaccines will be allocated first to other groups prioritized for mRNA vaccines, if any (none in this example). Subsequently, vaccines will be distributed to other age groups without prioritization, specifically adults and the elderly in this example.}
  \label{figure:stride_vaccines}
\end{figure}

Our setup includes two types of vaccines corresponding to those available in Belgium during the initial vaccination campaign \cite{Vaccines2024}: mRNA and vector-based vaccines \cite{Nagy2021}. The BNT162b2 vaccine by Pfizer-BioNTech and the mRNA-1273 by Moderna are grouped as mRNA vaccines. Analogously, AZD1222 by Oxford-AstraZeneca and Ad26COV2S by Janssen are both grouped as vector-based vaccines.
Each simulation day, the reported supply of mRNA and vector-based vaccines—corresponding to the actual doses delivered in Belgium since January 1, 2021 \citep{Vaesen2022}—is allocated to a selected group of individuals.
Weekly delivery quantities are extrapolated into daily vaccine uptakes assuming an uniform distribution over the days of a week. (Figure \ref{figure:vaccine_supplies}).

We define a vaccination strategy as a quintuple of the different vaccine types, \tc{relative to the 5 considered age groups}: Children (0-4), Youngsters (5-18), Young Adults (19-25), Adults (26-64) and Elderly (65+). The quintuple remains fixed throughout one simulation. 
When vaccinating the population according to a vaccination strategy, we randomly select unvaccinated individuals of the appropriate age groups to vaccinate. The number of vaccines per age group is specified based on the reported time-specific vaccine supply. This supply is proportionally distributed to the different age groups based on their respective sizes.
When all members of a particular age group are vaccinated, the vaccines will be divided among the other age groups, prioritizing age groups with the same vaccine type assigned.
This ensures that no vaccines are wasted in the simulation.
As an example, we present the vaccine administration for one of the evaluated strategies in Figure \ref{figure:vaccine_distribution}.

In each simulation, we target a vaccine uptake. Vaccination uptake is organized by household, so we randomly select households from the STRIDE population, until the target uptake levels are met. We refer to this random households selection as the \emph{uptake cohort}. During the simulation, vaccines will be allocated only to household members in the \emph{uptake cohort}, considering the age restrictions outlined in the vaccination strategy.

For the vaccine to reach its full protection, it requires some time after its administration, as neutralizing antibodies and virus-specific T cells must be produced \citep{Teijaro2021}.
We model this effect using a linear activation function, which linearly increases over a given time span, starting at the time when the first dose of the vaccine is administered.
We assume the vaccine's maximum efficacy is reached after 6 weeks, which mimics full vaccination scheme with a second dose after 4 weeks and maximum efficacy expected 2 weeks later \citep{CDC}.\footnote{For the Janssen vector-based vaccine, only one dose was administered. As these vector-based vaccines have a similar working mechanism, we assume the same activation function.}
We adopt differential vaccine efficacies for the Alpha VOC from the literature.
For the mRNA vaccines, we assume a vaccine efficacy $VE_S = 95\%$ for the susceptibility, $VE_I = 95\%$ for infectiousness and $VE_D = 100\%$ for the propensity to protect from severe disease \citep{Polack2020}. For the vector-based vaccines we assume $VE_S = 67\%$, $VE_I = 67\%$ and $VE_D = 100\%$ \citep{Knoll2021}.

\subsection{Disease outbreak outcomes}
\label{section:outcome_posteriors}
There are two possible outcomes for an infectious disease outbreak: either the disease spreads beyond a local context to become a fully established epidemic or it fades out \citep{Watts2005}.
Therefore, the distribution of the epidemic sizes is bimodal, which is reflected by most stochastic epidemiological models \citep{Watts2005}.
In the context of this study, where we consider an ongoing COVID-19 epidemic, we can focus on the mode of the infection size distribution that is associated with the established epidemic. 
This distribution is known to be approximately Gaussian \citep{Libin2018, Britton2010}.
We note that this argument does not automatically hold for the hospitalisation size distribution, as for many infectious diseases, the likelihood to be hospitalised is not uniform within a population \citep{Luk2001}. For COVID-19, hospitalisation rates rise exponentially with age \citep{Palmer2021}. Nonetheless, as for a particular scenario arm, we keep the contact reductions and vaccine policy constant, we still expect a central trend that can be well approximated with a Gaussian.

To incorporate this prior knowledge in BFTS, we consider the reward distribution Gaussian with unknown mean and variance and assume an uninformative Jeffreys prior $(\sigma)^{-3}$ on $(\mu, \sigma^2)$ \citep{Honda2014}. This prior leads to the non-standardised t-distributed posterior, that we truncate on the interval $[0,1]$ as we know the arm's means are in this interval. 
The formal derivation for this posterior can be found in \supplementary{A}.

\subsection{COVID-19 bandit}
\label{section:covid_experiments}

In the COVID-19 setting, we aim to find the prevention strategy that minimizes the proportion of the population affected at the end of the simulation (i.e., the 
attack rate, abbreviated as AR). This rate can be estimated with regards to infections (\infections) or hospitalisations (\hospitalisations).
To minimize the attack rate, we take the complement as a reward signal: $1-\infections$ for infections and $1-\hospitalisations$ for hospitalisations.

Given the 5 age groups and 2 vaccine types, with no vaccine as a third option, the bandit would have a total of $3^5$ arms.
In order not to waste any vaccines, we disregard all arms that do not use both types of vaccines, which results in a bandit with 180 arms.

While the bandit learns, it pulls an arm based on its sampling strategy. This arm is then translated to a corresponding vaccination strategy for each of the age groups. When pulling an arm, the bandit runs a STRIDE simulation for 4 calendar months, where the chosen vaccination strategy is executed until the end of the simulation.

\subsection{Establishing a ground truth to evaluate the framework}

\tc{During a pandemic, the efficient distribution of vaccines is crucial to reach the largest possible number of people.} However, in practice, vaccine uptake is often lower than expected, e.g., due to factors such as vaccine hesitancy \cite{Chaudhuri2022}.
Furthermore, the European Centre for Disease Prevention and Control has emphasized the need for interventions to boost vaccine uptake to effectively control COVID-19 \cite{Uptake2024}.
In Belgium, a vaccine uptake rate of 75.7\% was recorded later in 2021 \cite{Catteau2021}, and accordingly, we adopt a vaccine uptake proportion of 75\% in our simulations.

\begin{figure}[h!]\centering
  \includegraphics[width=\linewidth]{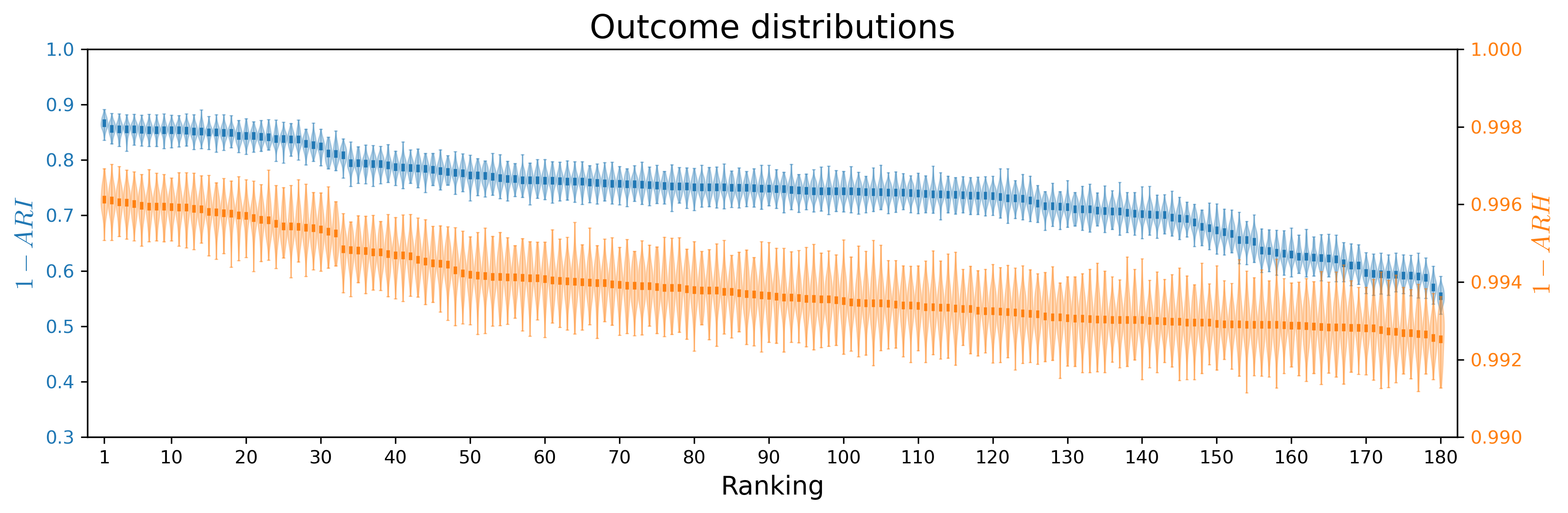}
  \caption{Ground truth of all arms for the baseline scenario, ranked based on 100 stochastic simulations. Ground truth of all arms for the baseline scenario, ranked based on infections (1 - \infections) and hospitalisations (1 - \hospitalisations). We note that the ranking based on infections versus hospitalisations does not necessarily match, here we show an independent ranking for each of the criteria.}
  \label{figure:config0_all_arms}
\end{figure}

To validate our method, we establish a baseline scenario with a 75\% uptake proportion (Table \ref{table:configs}), where we obtain 100 simulation replicates for each of the strategies, using the STRIDE stochastic individual-based model. This ground truth will be used to assess the performance of the algorithms to identify the true set of optimal arms.
Figure \ref{figure:config0_all_arms} shows the reward distributions for 100 simulation replicates of each of the arms.

\begin{figure}[h]
    \centering
    \includegraphics[width=0.65\linewidth, keepaspectratio]{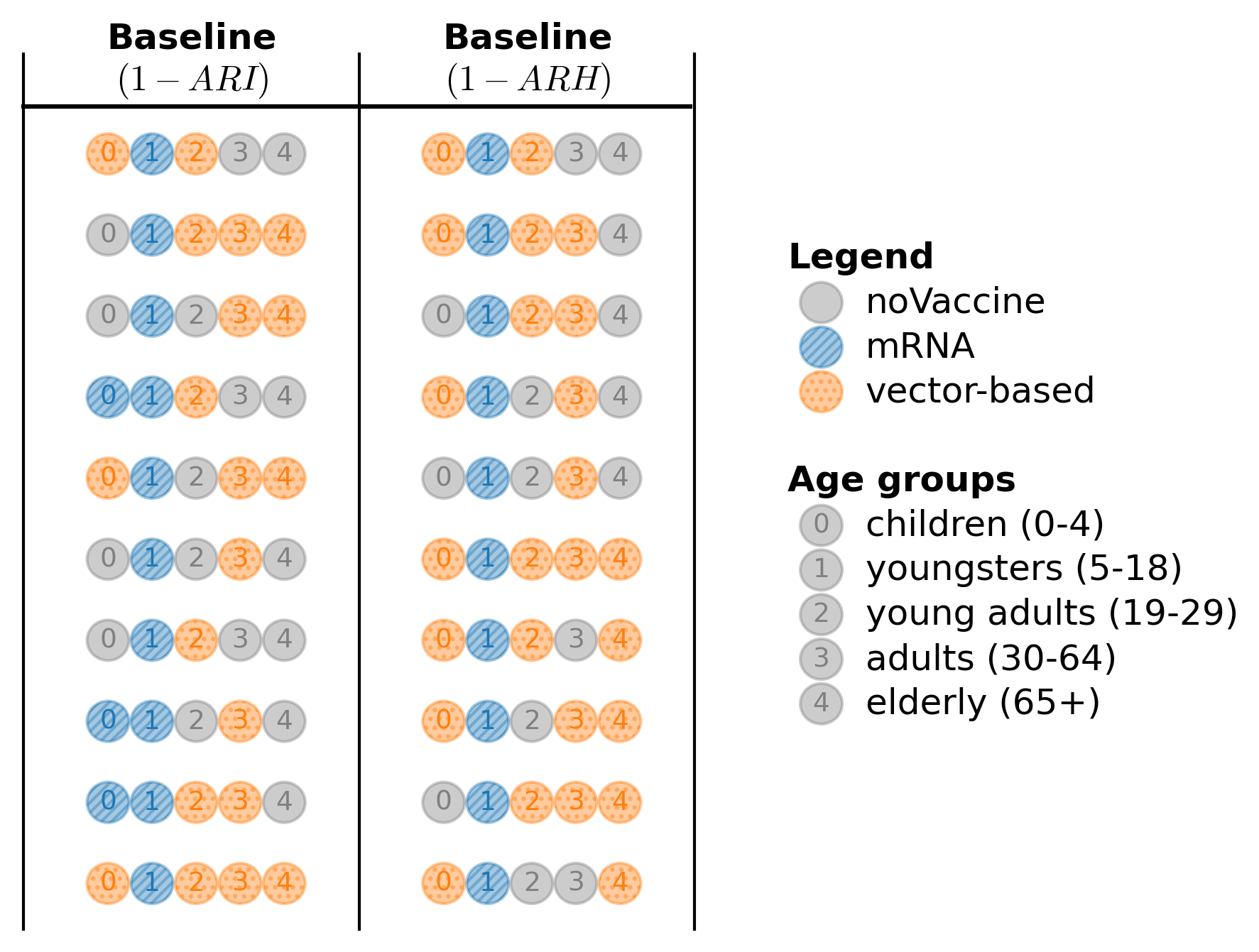}
    \caption{Ground truth of the top-$10$ vaccination strategies for the baseline scenario when minimising the infection (\infections) and hospitalisation (\hospitalisations) attack rates. Each strategy in the top-$10$ strategies is represented by 5 numbered circles, each representing a specific age group as highlighted in the legend. The colour of the circle indicates which vaccine type is being prioritised for the given strategy. For example, the first strategy when optimising for \infections\ prioritises \adeno\ vaccines for children and young adults. Youngsters receive priority for an \mRNA\ vaccine, while adults and elderly receive no vaccine priority.}
    \label{figure:config_0_ground_truth}
\end{figure}

The true $m$-top vaccination strategies demonstrate a distinct trend for the infection attack rate \infections\ and the hospitalisation attack rate \hospitalisations\ (Figure \ref{figure:config_0_ground_truth}). Most noticeably, all 10 top strategies prioritise vaccinating youngsters with \mRNA\ vaccines. 
\tc{Children do not receive a particular recommendation in the top-$10$ strategies for \infections, as all vaccine types, including no vaccine, are present in these top-$10$ strategies. This indicates that assigning a particular vaccine type priority to children is less critical when reducing infections.}
When optimizing for \hospitalisations, children are prioritized and receive \adeno\ vaccines in 7 of the top strategies.
Young adults, adults and elderly receive \adeno\ vaccines if they are prioritised. Any remaining vaccines will be distributed among the remainder of the unvaccinated population. As a result, all age groups will eventually be vaccinated once the target age groups have been covered. 
We observe some overlap between the strategies for \infections\ and \hospitalisations, which is expected since reducing overall infections also contributes to lowering hospitalizations.

Using this ground truth, we compare the performance of BFTS, AT-LUCB and Uniform sampling. Uniform sampling aims to pull each arm an equal number of times by pulling the least-sampled arm at each timestep. Consequently, uniform sampling recommends the empirical $m$-top arms. Both the BFTS and AT-LUCB algorithm are described in Section \ref{sec:best_arm_id}.
We report the algorithms' performances using 
two statistics \citep{jun2016}.
The first statistic is the proportion of correctly recommended arms at time $t$,
\begin{equation}
    \label{eq:proportion_true_arms}
    \frac{|J^{(t)} \cap J^{\text{True}}|}{m}.
\end{equation}
$J^{\text{True}}$ denotes the true set of optimal arms, which we know via our ground truth, and $J^{(t)}$ denotes the set of recommended arms at time $t$.
The second statistic is the sum of the means of the $m$-top arms at time $t$,
\begin{equation}
    \label{eq:sum_of_means}
    \sum_{a \in J^{(t)}} \mu_a.
\end{equation} 

\begin{figure}[H]\centering
  \begin{subfigure}[h!]{0.495\linewidth}\centering
    \includegraphics[width=\linewidth]{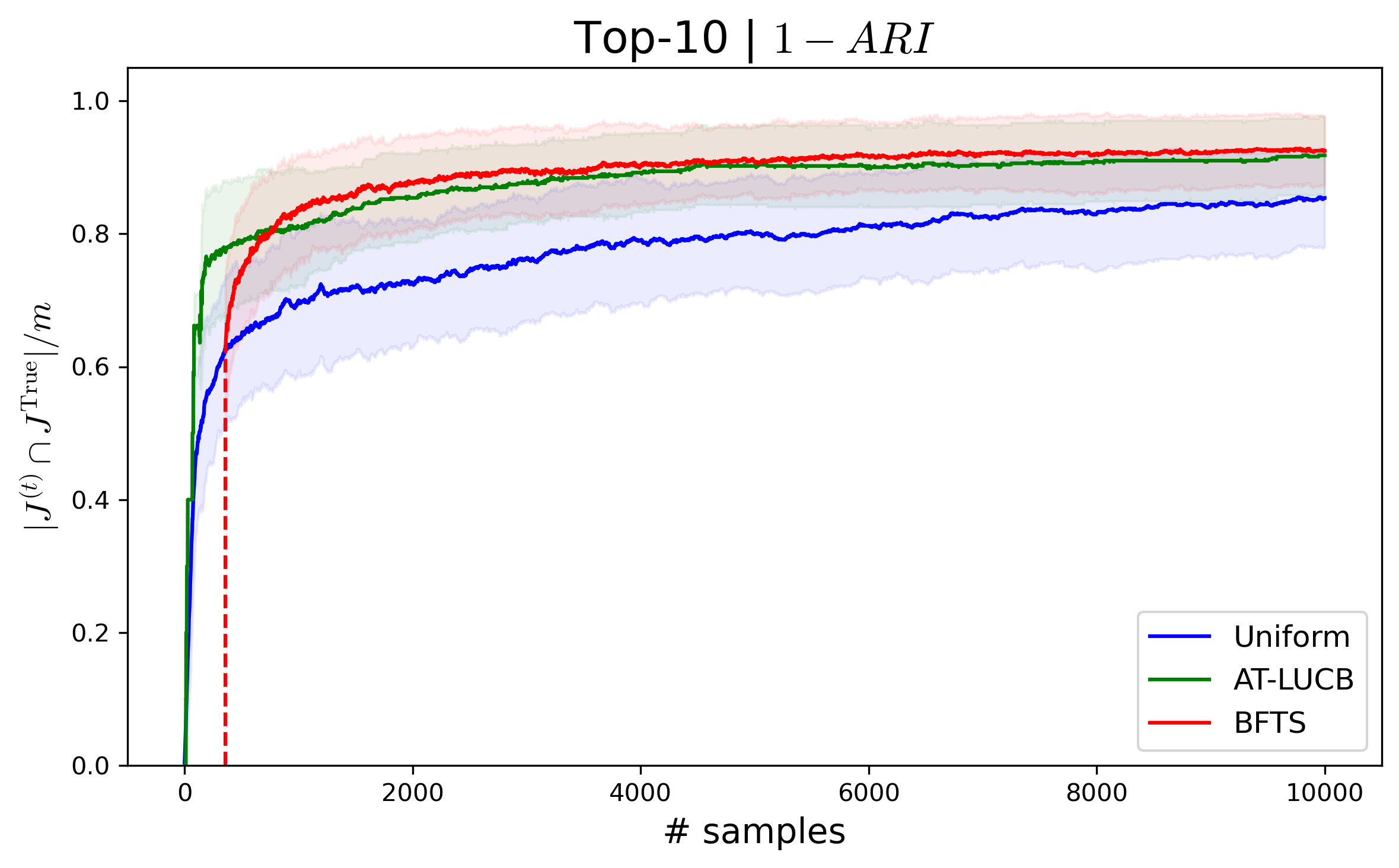}
  \end{subfigure}
    \begin{subfigure}[h!]{0.495\linewidth}\centering
    \includegraphics[width=\linewidth]{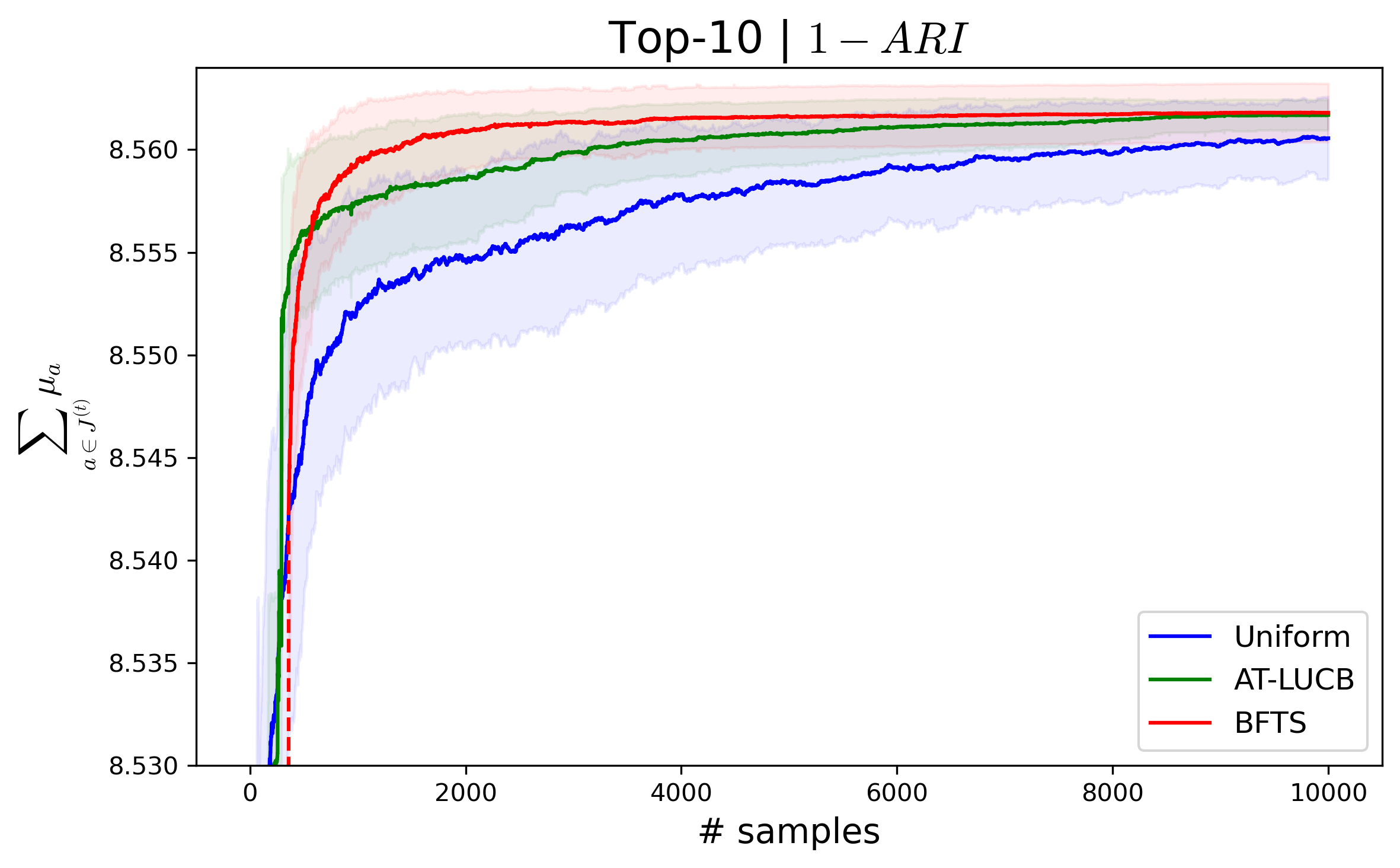}
  \end{subfigure}
  \begin{subfigure}[h!]{0.495\linewidth}\centering
    \includegraphics[width=\linewidth]{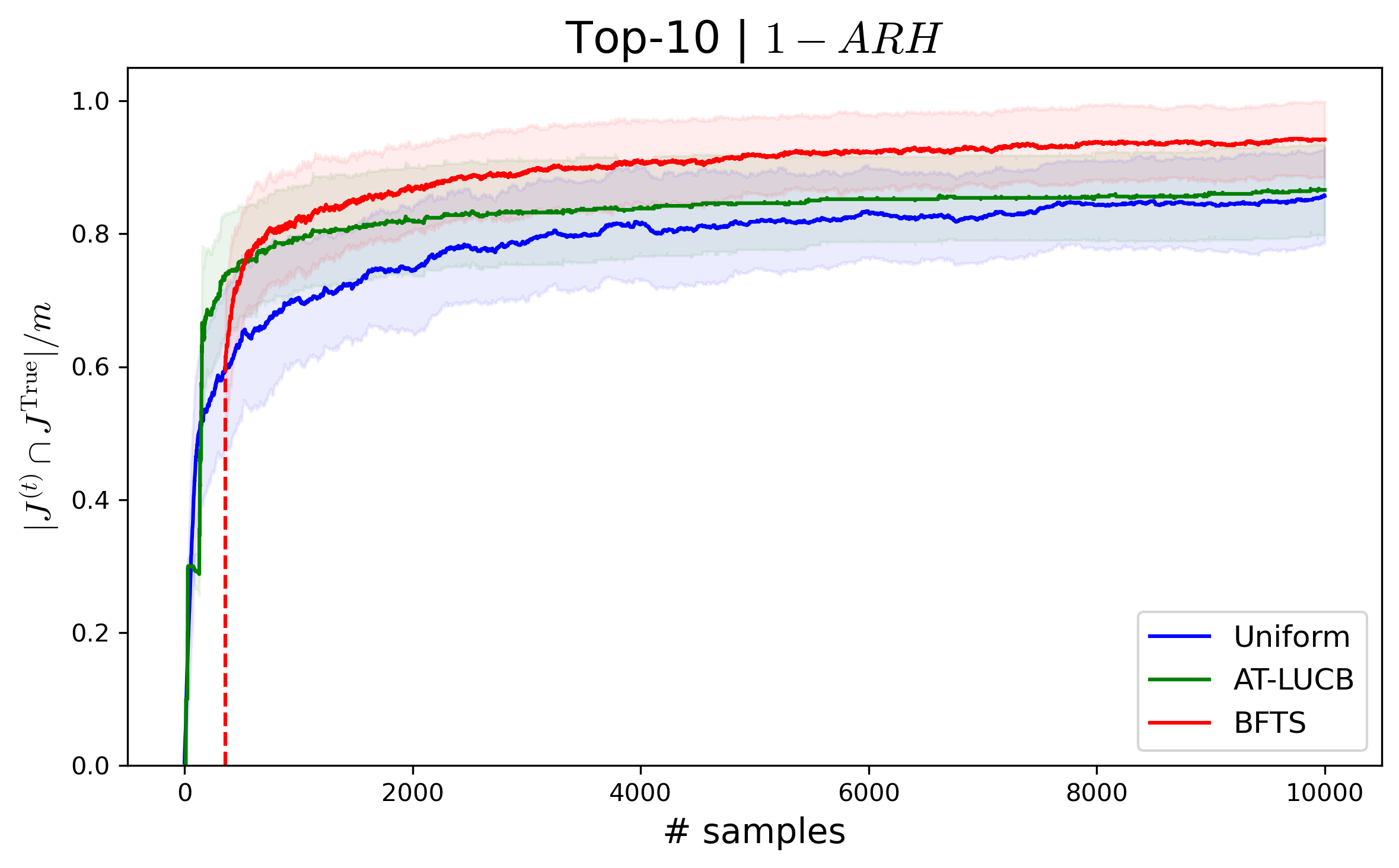}
  \end{subfigure}
    \begin{subfigure}[h!]{0.495\linewidth}\centering
    \includegraphics[width=\linewidth]{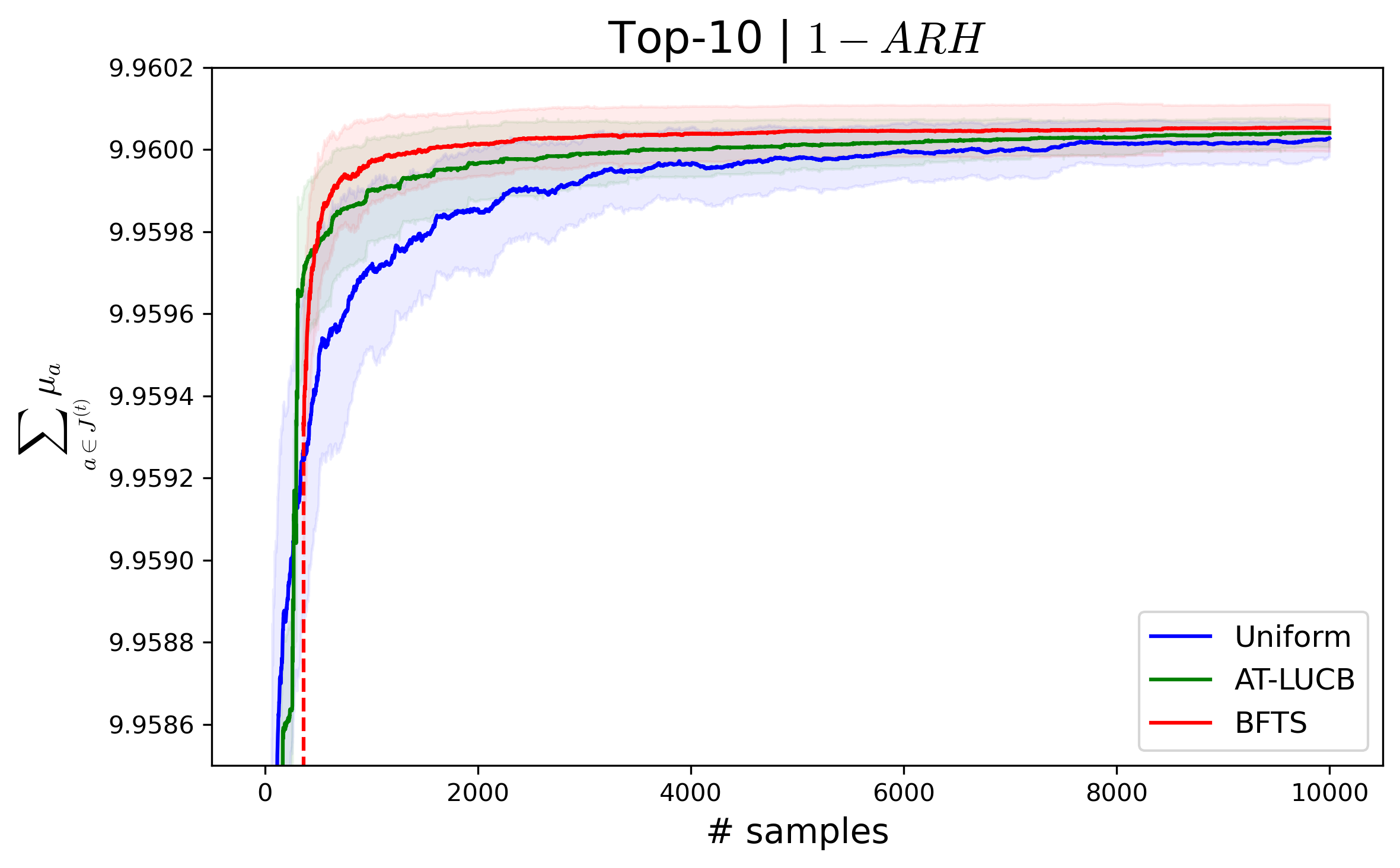}
  \end{subfigure}
  \caption{Learning curves for the ground truth based on infections (\infections) and hospitalisations (\hospitalisations), top vs bottom row, respectively. Left column: The average proportions of correctly ranked arms, with standard deviation. Right column: The average sum of true means, with standard deviation.}
    \label{figure:gt_inf_hosp_results}
\end{figure}

Note that uniform sampling and BFTS obtain one sample per time step, whereas AT-LUCB samples twice per timestep. We plot the results in terms of the number of samples (x-axis in Figure \ref{figure:gt_inf_hosp_results}) to facilitate a fair comparison.
We consider truncated t-distribution posteriors for BFTS (as detailed in Section \ref{section:outcome_posteriors}).
Figure \ref{figure:gt_inf_hosp_results} shows the results of 100 simulation replicates per algorithm, over 10,000 samples, measured in terms of infections and hospitalizations.
To obtain a proper posterior for BFTS, each arm's posterior needs to be initialised twice \citep{Honda2014}. In general, BFTS needs this short period to meet AT-LUCB's performance, but quickly outperforms AT-LUCB after this warm-up period.


\begin{figure}[H]\centering
  \begin{subfigure}[h!]{0.495\linewidth}\centering
    \includegraphics[width=\linewidth]{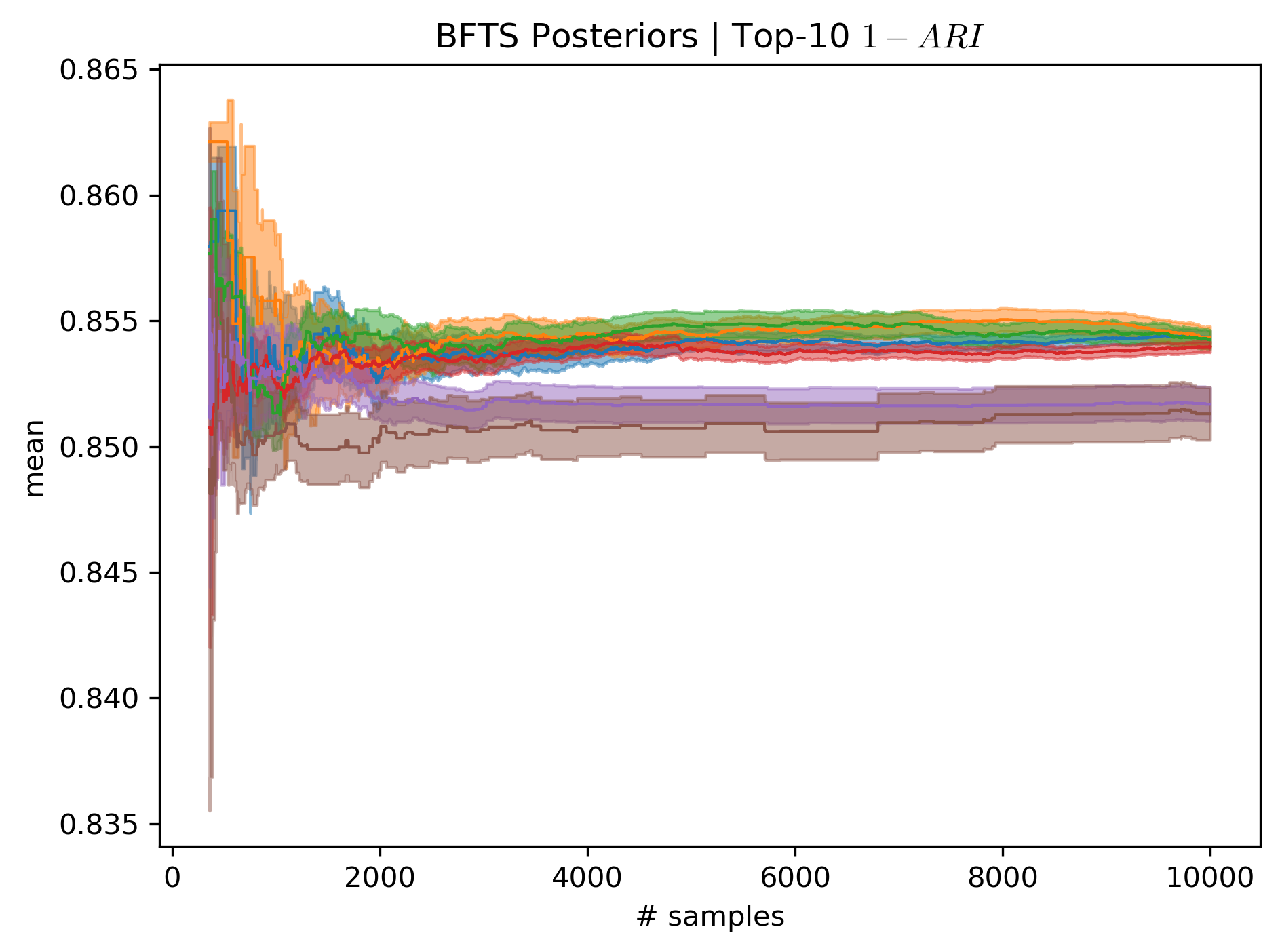}
    \caption{}
  \end{subfigure}
    \begin{subfigure}[h!]{0.495\linewidth}\centering
    \includegraphics[width=\linewidth]{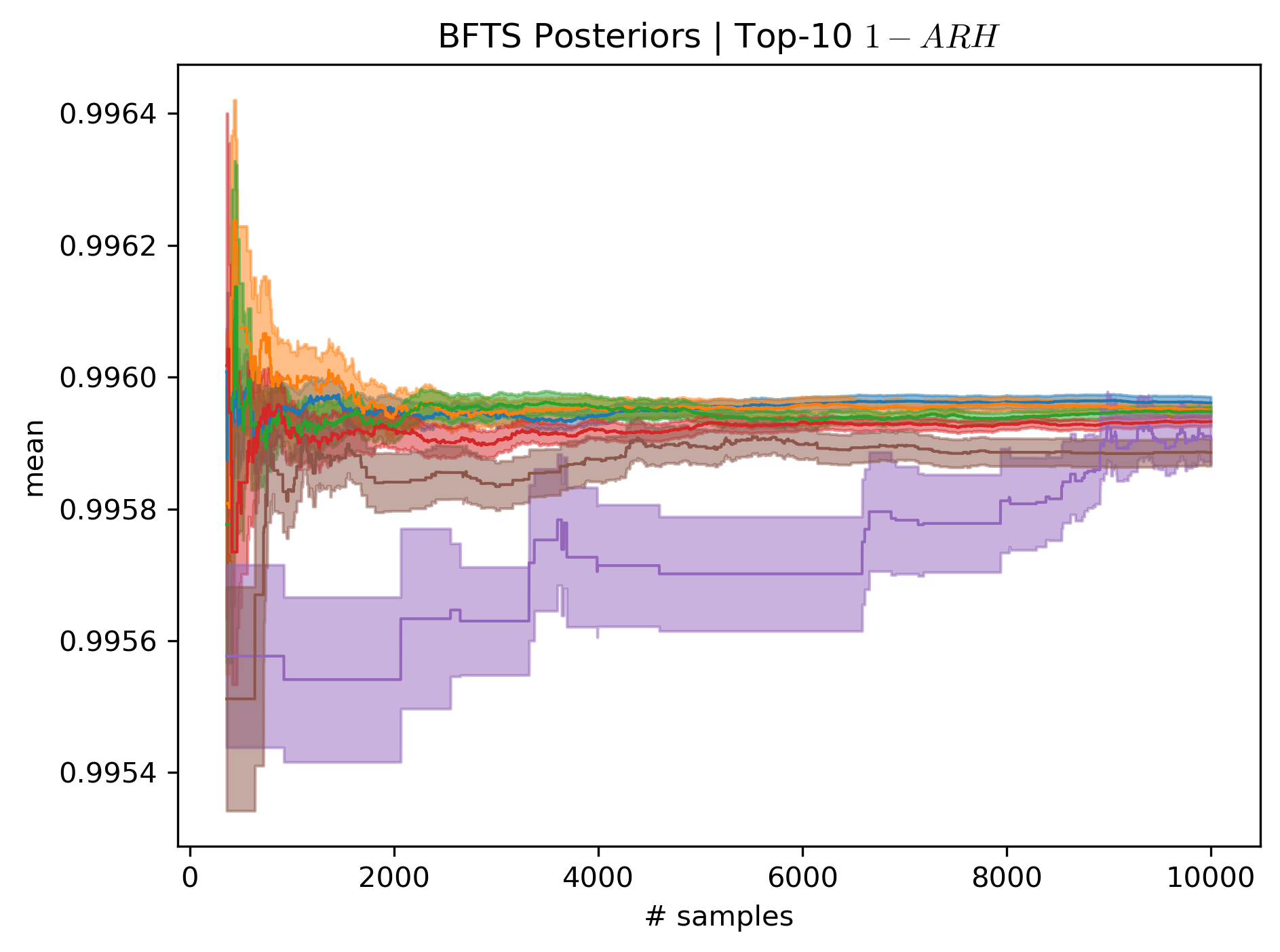}
    \caption{}
  \end{subfigure}
  \caption{Posteriors for the \scenarioZero\ scenario concerning (a) infections and (b) hospitalisations. The estimated means and uncertainties (standard deviations) are shown for the 3 arms above and the 3 arms below the decision boundary. Note that the arms closest to the decision boundary have a reduced uncertainty, as the bandit focused on these arms to reduce its uncertainty about the decision boundary.}
    \label{figure:stride_posteriors_a}
\end{figure}

Figure \ref{figure:stride_posteriors_a} shows the posteriors' estimated means and uncertainty (standard deviation) for the 3 arms above and the 3 arms below the decision boundary of a single bandit run of BFTS.
As BFTS pulls an arm, it reduces its uncertainty for that arm. However, as the true means are close to each other (see Figure~\ref{figure:config0_all_arms}), there still remains uncertainty with regards to the estimated top-$10$ arms.
We note that inspecting how these posteriors evolve over time presents an interesting way for decision makers to interpret and report the algorithm's recommendations and the uncertainty associated with these recommendations.

\subsection{Analysing vaccination policies under various contact reduction schemes}

We define a 180-armed bandit to learn the top-$10$ vaccination strategies using BFTS, for the different contact reduction scenarios mentioned above (Table~\ref{table:configs}). 
Due to the computational burden of the STRIDE model accounting for the 11 million population for Belgium, running a single simulation, that is optimised and multi-threaded, takes approximately 5-6 minutes on the Genius and Hydra Vlaams Computer Centrum (\url{https://www.vscentrum.be}) high performance computing infrastructure, for our configurations. Each time the bandit pulls an arm, a new simulation is run. Consequently, the time required to perform experiments increases quickly due to the sequential nature of the bandit setting.
As a result, we have set a limit of 2000 simulations (i.e., arm pulls) per experiment to obtain results equivalent to a uniform evaluation of 18,000 simulations. The 2000 simulations already correspond to about 1.5 weeks of computation on the Genius VSC high performance computing infrastructure. In the discussion section, we view further scaling of the simulations as a direction for future work.

For the baseline COVID-19 scenario we obtained a ground truth to validate our results.
For the other social contact scenarios, we evaluate our bandit framework for what it was intended: to find the top strategies when evaluating each arm is computationally unfeasible. Therefore, as we do not have a ground truth, we evaluate the quality of the obtained results by investigating the bandit's decision uncertainty about the learned top vaccination strategies.

In this analysis, we investigate vaccine allocation strategies under distinct contact reduction policies. The \scenarioOne\ scenario (Table \ref{table:configs}) imposes $50\%$ contact reductions in the workplace and community, with primary schools open at full capacity.
Secondary schools operate at 50\% capacity, while universities and colleges are closed \citep{Willem2021}. 
This scenario considers moderate restrictions at work and in the community, requiring a well-chosen vaccination strategy to counteract the additional contacts compared to the baseline scenario.
In the \scenarioTwo\ scenario, we follow the same contact reductions as the baseline, with the exception of having tertiary schools open at $70\%$ contact reductions.
The \scenarioThree\ scenario explores the case where primary, secondary schools and universities are open. As the baseline scenario has shown that children should be prioritised when vaccinating, this scenario provides an interesting perspective as school contacts for children and youngsters are fully allowed.
The \scenarioFour\ and \scenarioFive\ scenarios both consider a middle-ground between the baseline and relaxed scenario.
The uncertainty analysis (based on the analysis of the posteriors) for all these scenarios can be found in \supplementary{D}.

\begin{figure}[H]
    \centering
    \includegraphics[width=1.0\linewidth, keepaspectratio]{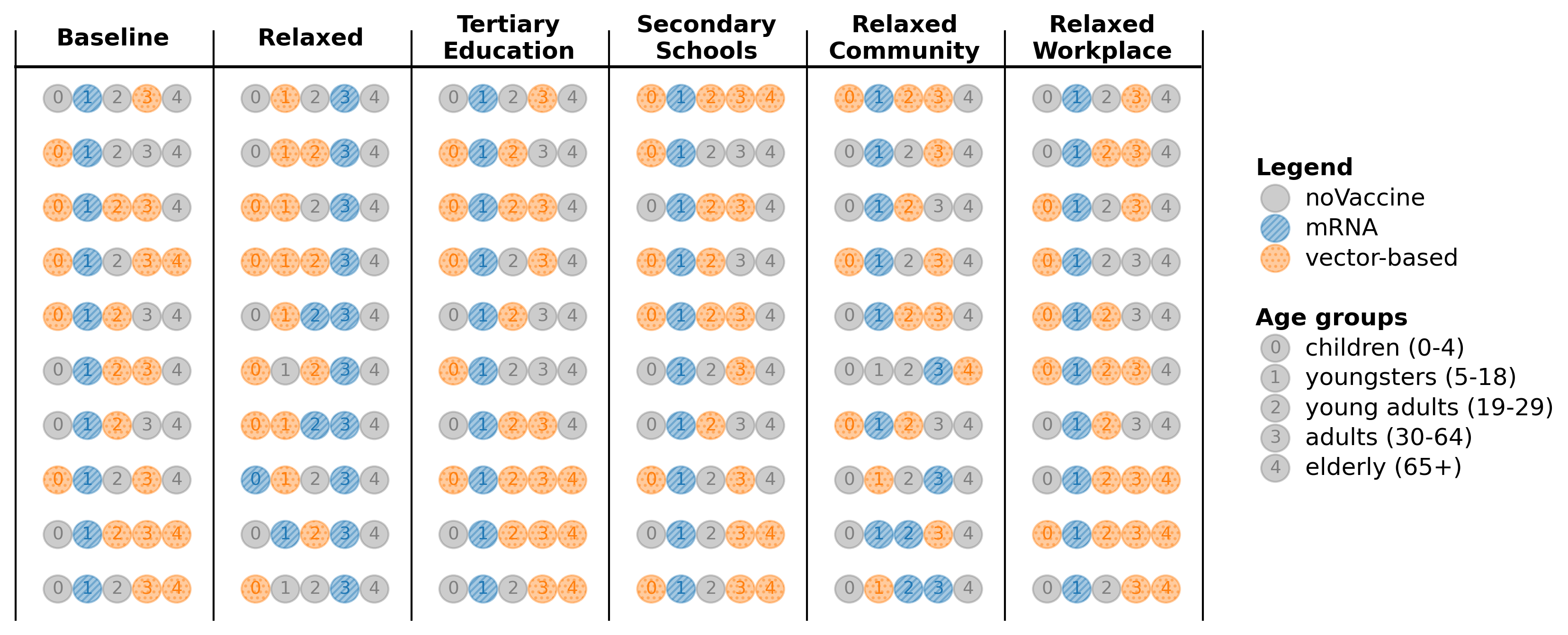}
    \caption{Learned top-$10$ vaccination strategies when minimising the infection attack rate (\infections) under various contact reduction schemes, under a 75\% vaccine uptake proportion.}
    \label{figure:config_rankings_inf0.75}
\end{figure}

When minimising the infection attack rate (\infections), there is a clear trend regarding youngsters across the scenarios (Figure \ref{figure:config_rankings_inf0.75}). In all scenarios except the \scenarioOne\ and \scenarioFour\ scenarios, the top vaccination strategies exclusively prioritise \mRNA\ vaccines for youngsters. In contrast, in the \scenarioOne\ scenario, our analysis recommends to prioritise adults with \mRNA\ vaccines. As reducing infections is the priority, the most rewarding strategies are those that prioritise youngsters, young adults and adults as they are making more contacts compared to the \scenarioZero\ scenario. 
It is worth noting that prioritisation of vaccines for the elderly seems linked to the social contact scenario, suggesting that the social contacts made by the elderly may have a greater impact on the infection attack rate, in these scenarios.

\begin{figure}[H]
    \centering
    \includegraphics[width=1.0\linewidth, keepaspectratio]{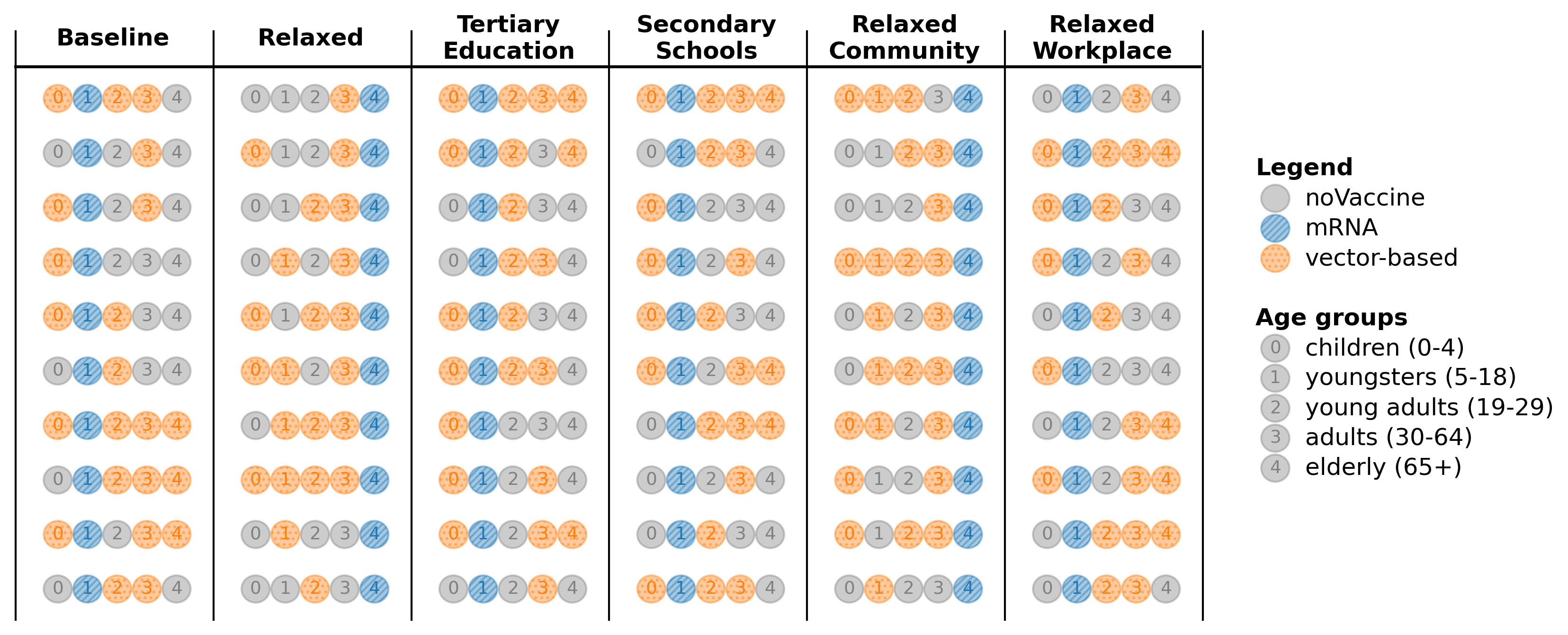}
    \caption{Learned top-$10$ vaccination strategies when minimising the hospitalisation attack rate (\hospitalisations) under various contact reduction schemes, under a 75\% vaccine uptake proportion.}
    \label{figure:config_rankings_hosp0.75}
\end{figure}

For the hospitalisation attack rate (\hospitalisations), we notice that youngsters are still prioritised with \mRNA\ vaccines. However, there is a shift in focus for the \scenarioOne\ and \scenarioFour\ scenarios, where all top-$10$ strategies prioritise giving elderly \mRNA\ vaccines (Figure \ref{figure:config_rankings_hosp0.75}). Both scenarios  allow more community contacts, where there is greater involvement of the elderly compared to school or work activities. As the older population is more likely to be hospitalised \citep{Palmer2021}, the bandit learns to vaccinate them first. Both vaccine types have the same efficacy in preventing severe disease and hospitalizations. However, \mRNA\ vaccines are more effective in reducing susceptibility and infectiousness. Combined with their greater availability during the simulation, this makes them the preferred option for vaccinating and protecting the elderly from hospitalization.

Interestingly, these results indicate for each contact reduction scenario which age group should be prioritised for vaccination. As the supply of \mRNA\ vaccines is higher than the supply of \adeno\ vaccines throughout the experiments, choosing \mRNA\ for an age group means that this age group will receive a majority of the vaccines, thereby reducing that particular age group's impact on the attack rate. We refer back to Figure \ref{figure:stride_vaccines} for an example of this prioritisation effect based on the supply.
For example, in the \scenarioZero\ scenario for \hospitalisations, even though hospitalisations were a priority, only three arms prioritise vaccinating the elderly because the impact of other age groups appears to be more important (Figure \ref{figure:config_0_ground_truth}).
Similarly, the presence of multiple vaccine types for an age group in the top-$10$ strategies suggests that the specific vaccine type is less critical for that particular age group, as long as the individuals in this age group are vaccinated. For example, the \scenarioZero\ scenario for \infections\ (Figure \ref{figure:config_0_ground_truth}) indicates less importance regarding the vaccination type when it comes to children, as \mRNA\ and vector-based vaccines were both recommended options.
We conclude that these results are heavily influenced by the effectiveness of the vaccines in significantly reducing transmission likelihood. With the emergence of new variants, these odds changed, which we reflect upon in the discussion section.

\begin{figure}[H]
    \centering
    \includegraphics[width=1.0\linewidth, keepaspectratio]{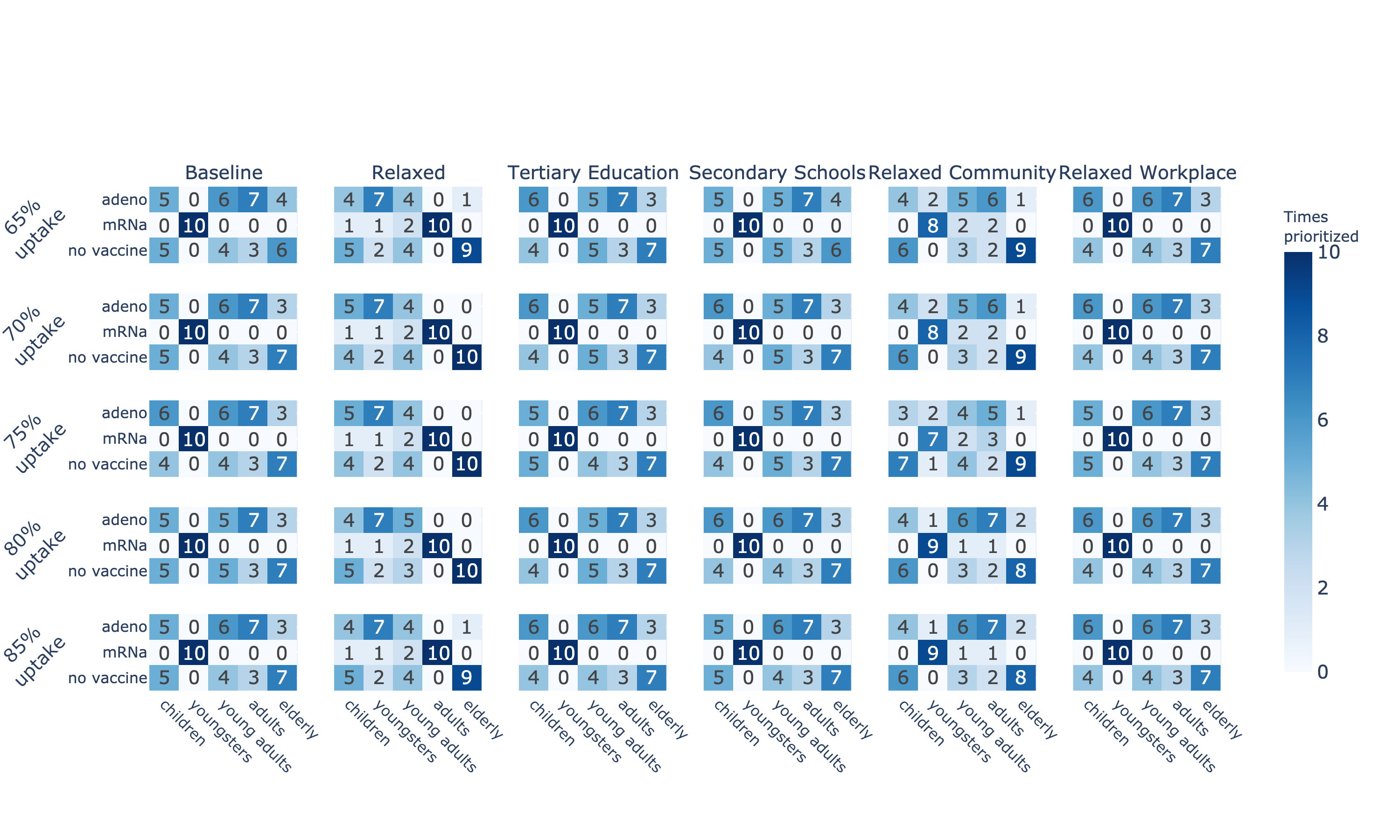}
    \caption{Priorities of vaccine types per age group across all uptakes, when optimising the infection attack rate (\infections) under various contact reduction schemes and uptake proportions.}
    \label{figure:config_rankings_inf_heatmap}
\end{figure}
\begin{figure}[H]
    \centering
    \includegraphics[width=1.0\linewidth, keepaspectratio]{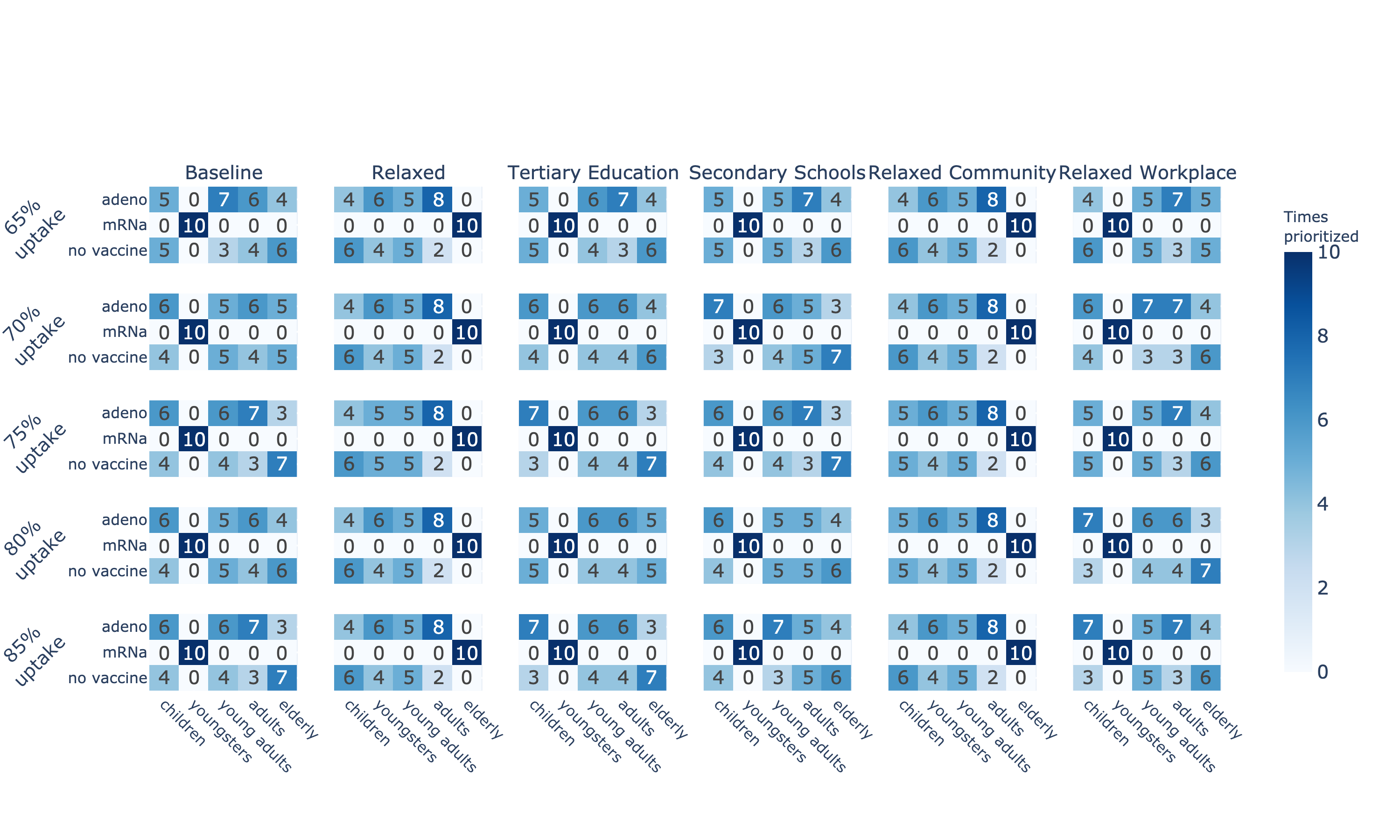}
    \caption{Priorities of vaccine types per age group across all uptake proportions, when optimising the hospitalisation attack rate (\hospitalisations) under various contact reduction schemes and uptake proportions.}
    \label{figure:config_rankings_hosp_heatmap}
\end{figure}

To summarise these results, in Figure \ref{figure:config_rankings_inf_heatmap} we show an overview of how often each age group is prioritised for a given vaccine type in the learned top-$10$ arms, under distinct vaccine uptake proportions. We note that across different uptake proportions, the number of times an age group is prioritised (for a given vaccine type) remains similar. While the overall age group priorities are similar for a given scenario, the best strategies learned might differ in their vaccine type combinations across age groups. Figure \ref{figure:config_rankings_hosp_heatmap} shows an overview of the prioritisation when \hospitalisations\ is optimised. Here we observe similar priorities within a contact reduction scenario across the different uptake proportions. 
We present additional results on the specific top-$10$ strategies for all contact reduction scenarios, for \infections\ and \hospitalisations, in \supplementary{E}. Moreover, we show results for different uptake proportion of 65\%, 70\%, 80\% and 85\% in \supplementary{E}.

\section{Discussion}

In this work, we present a multi-armed bandit framework to study mitigation policies in individual-based epidemiological models. With this framework, we study vaccine allocation policies in a COVID-19 epidemic. Via a ground truth analysis, we show it is possible to efficiently learn the best strategies using a limited number of stochastic model evaluations. Additionally, the bandit allows policy makers to use the learned posteriors and their uncertainty to make informed decisions. Through our vaccine allocation study, we highlight the connection between targeted social contact reductions and the design of vaccine allocation policies.

\tc{Through our framework, we present a comprehensive retrospective analysis on the organisation of COVID-19 vaccination policies under different contact reduction schemes. Moreover, we investigate the impact of vaccine uptake proportions. Through our experiments, we show that the top policies follow a clear trend regarding the prioritised age groups and assigned vaccine type, which provides insights for future vaccination campaigns. When varying the overall uptake levels, our experiments suggest that this has limited influence on the optimal vaccine policy design.} 
Next to providing retrospective insights regarding COVID-19 vaccine allocations, we contribute a free software (GPL-licensed) framework that will facilitate the investigation of mitigation policies for future pandemics.

For simplicity, we make certain modeling assumptions. First, we keep contact reductions constant during a simulation. Second, we do not consider imported cases from abroad, motivated by the fact that we consider an ongoing epidemic that is mainly driven by intra-country contact dynamics. 
This study concerns a retrospective analysis that assumes that vaccine delivery dates are known. To reason about policies when the delivery scheme of vaccines is uncertain, we consider the use of stateful reinforcement learning for future work.

We note that the vaccine efficacies are representative for the start of the vaccination campaign, but as variants continued to emerge, vaccine efficacy regarding susceptibility and infectiousness has decreased significantly. Nonetheless, our study provides insights to optimal policies at the start of the vaccination campaign. We consider the evaluation of vaccination policies under the emergence of distinct VoC as future work. 
Additionally, we consider all age groups in the vaccine allocation study. However, as for the SARS-CoV-2 pandemic new vaccine platforms were trialed, these vaccines were only approved for 18 years and older at the start of vaccination campaign \citep{WHOc}.
 Our analyses do show that a rapid adoption of vaccines by children and/or youngsters could have an important impact on the epidemic and could lead to lower contact reductions. This is an important consideration for future vaccination campaigns, which might target children earlier on, as the mRNA and vector-based vaccine platforms have now undergone rigorous evaluation \citep{Baden2024, ElSahly2021}.
%
In the context of COVID-19, it was shown that individuals might increase their contacts once they have been vaccinated \citep{Wambua2022}. We consider such behavioral aspects an interesting aspect to study in future work. Furthermore, as we consider a limited time period (4 months), vaccine waning is not considered in this study \cite{Ferdinands2022}. We do acknowledge that this would be interesting to consider for future work, when evaluating long term mitigation policies. Moreover, we consider it an interesting venue for future work to consider robustness of vaccination policies with respect to the emergence of variants.

In this work, we do not explicitly consider correlations between arms, to establish a generic policy evaluation framework that supports decision uncertainty. We note that the Bayesian exploration scheme will implicitly account for such correlations. While there are bandit algorithms that can exploit such correlations \citep{Wang2018,Gupta2021,Singh2020}, to the best of our knowledge there exist no Bayesian $m$-top exploration algorithms, which thus constitutes an interesting direction for future work.

Running sequential simulations for the bandit algorithm on STRIDE increases the time needed to conduct experiments. Therefore, our bandits framework would strongly benefit from parallelisation with regards to the pulled arms. We note that an extension of the Bayesian $m$-top algorithm with a delayed bandit approach constitutes an important venue for subsequent work \citep{Gael2020}. Furthermore, additional optimisations and parallelisation to reduce the execution time of a single STRIDE simulation even more, could be explored.

In addition to the vaccine allocation policies studied here, our approach can be extended to explore other diseases and mitigation strategies for both disease and transmission, such as antiviral allocation strategies \citep{Torneri2020}.
From an epidemiological perspective, future work may focus on the impact of universal testing approaches to mitigate the epidemic \citep{Libin2021b}, and repetitive testing in a school environment \cite{Leung2022}. Similarly, the effect of superspreading \citep{Kuylen2022} and its impact on social distancing and vaccine allocation presents interesting venues for future research.
From a fairness perspective, future work may take multiple fairness requirements into account next to minimising infections or hospitalisations, by considering a multi-objective approach \citep{Cimpean2024}.

While our analysis reveals interesting insights in vaccine allocation strategies, it is important to note that these policies were learned within the limitations of the model used. To use the policies in a real epidemic emergency, a thorough validation is warranted.

For the vaccine allocation analyses under different contact reductions, we only allow the bandit a budget of 2000 samples, due to the computational burden of these analyses. On the one hand, this leaves room for uncertainty on the decision boundary, as was shown in our ground truth analysis. On the other hand, our ground truth analysis showed that BFTS is able to achieve good performance after 2000 steps, and our inspection of the posteriors of the contact reduction analyses confirmed this. We do stress that inspecting the posteriors is important, and when a high uncertainty is observed, this might warrant additional steps. This is possible, using the anytime framework we present.

Our study assumes that household-based clustering of vaccine uptake serves as a reliable proxy for vaccine hesitancy. While supported by prior research \cite{Chaudhuri2022}, this assumption warrants further scrutiny. Households capture collective decision-making tendencies, but it may overlook individual variability. Adolescents, for instance, often exhibit greater autonomy in health-related decisions compared to younger children \cite{Fazel2021,Yang2019}. Additionally, external influences such as peer pressure, workplace mandates, or targeted public health campaigns may shape individual attitudes beyond household-level norms \cite{Betsch2015}.
To address these concerns, future work could involve accounting for demographic factors like age, education, and socioeconomic status that moderate decision-making within households. Moreover, alternative predictors such as geographic clustering or community-level factors could complement household-based analyses, offering a more nuanced understanding of vaccine hesitancy \cite{Brewer2017}.

\section*{Acknowledgments}
A.C. is funded by the Fonds voor Wetenschappelijk Onderzoek (FWO) via fellowship 1SF7823N and received funding from the Research Council of the Vrije Universiteit Brussel (OZR-VUB) through OZR mandate OZR3819. A.C. also acknowledges funding from the FWO COVID-19 research project G0H0420N.
All experiments were run on the Genius and Hydra clusters of the Flemish Supercomputer Center (Vlaams Supercomputer Centrum - VSC).
This work also received funding from the European Research Council (ERC) under the European Union’s Horizon 2020 research and innovation program (grant number 101003688 – EpiPose project).
P.J.K.L. gratefully acknowledges support from FWO via postdoctoral fellowship 1242021N and the Research council of the Vrije Universiteit Brussel (OZR-VUB via grant number OZR3863BOF).
N.H. acknowledges support from the Scientific Chair of Evidence-based Vaccinology under the umbrella of the Methusalem framework at the University of Antwerp.
N.H. and A.N. acknowledge funding from the iBOF DESCARTES project (reference: iBOF-21-027).
P.J.K.L. and L.W. acknowledge support from FWO grant G059423N.
L.W. gratefully acknowledges support from FWO postdoctoral fellowship 1234620N.
This research acknowledges funding from the Flemish Government through the AI Research Program.
This project was supported by the VERDI project (101045989), funded by the European Union. Views and opinions expressed are however those of the author(s) only and do not necessarily reflect those of the European Union or the Health and Digital Executive Agency. Neither the European Union nor the granting authority can be held responsible for them.
The funders had no role in study design, data collection and analysis, decision to publish, or preparation of the manuscript.

\section*{Reproducibility}
The code for the bandit framework, the $m$-top algorithms and the COVID-19 experiments that were conducted in this paper is available at \url{https://github.com/icimpean/m-top-covid}. The vaccine extension to the STRIDE simulator is available at \url{https://github.com/icimpean/stride/tree/vaccine}.

\section*{Use of generative AI}
The use of generative AI was limited to supporting roles, such as language refinement or idea generation, without replacing critical analysis, original research, or authorship contributions.

\newpage

\bibliography{bibliography.bib}{}
\bibliographystyle{plain}

\newpage
\begin{appendices}

\section{Truncated t-distribution}
\label{supplemental:truncated_t_distr}

We consider a Gaussian reward distribution with unknown variance and assume an uninformative Jeffreys prior $(\sigma)^{-3}$ on $(\mu, \sigma^2)$. 
Given rewards $\textbf{r}=\{r_1,...,r_n\}$, this prior leads to the non-standardised t-distributed posterior, that we truncate given that we know that the arms' means are in $[0,1]$:
\begin{equation}
  \label{eq:scaled_gaussian_posterior}
\mu \sim \mathcal{T}_{n,[0,1]}\left(\mu_0 =\frac{\sum_{i=1}^{n}r_i}{n}, \sigma^2_0 = \frac{\sum_{i=1}^{n}(r_i-\mu_0)^2}{n^2}\right).
\end{equation}

Given the pdf $f(\cdot)$ of a non-standardised t-distribution $\mathcal{T}_\upsilon(\mu,\sigma^2)$
\begin{equation}
\begin{split}
f(x) =\frac{\Gamma\left(\frac{\upsilon +1}{2}\right)}{\sigma \sqrt{\upsilon \pi} \Gamma\left(\frac{\upsilon}{2}\right)}\left(1 + \frac{(x - \mu)^2}{\upsilon \sigma^2}\right)^{-\frac{\upsilon + 1}{2}}
\end{split},
\end{equation}
and cdf $F(\cdot)$, we can compute the mean of the truncated non-standardised t-distribution using this normalised definite integral:
\begin{equation}
\label{eq:truncated_t_mean_base}
\frac{\int^1_0 x  f(x) dx}{F(\frac{1-\mu}{\sigma}) - F(\frac{0-\mu}{\sigma})}
\end{equation}
From this, we can derive an analytic expression by first considering the numerator:
\begin{equation}
\begin{split}
&\int^1_0 x f(x) dx\\
&=\int^1_0 x \frac{\Gamma\left(\frac{\upsilon +1}{2}\right)}{\sigma \sqrt{\upsilon \pi} \Gamma\left(\frac{\upsilon}{2}\right)}\left(1 + \frac{(x - \mu)^2}{\upsilon \sigma^2}\right)^{-\frac{\upsilon + 1}{2}} dx\\
&= \int^{x=1}_{x=0} \sigma\frac{x - \mu + \mu}{\sigma} \frac{\Gamma\left(\frac{\upsilon +1}{2}\right)}{\sqrt{\upsilon \pi} \Gamma\left(\frac{\upsilon}{2}\right)}\left(1 + \frac{(x - \mu)^2}{\upsilon \sigma^2}\right)^{-\frac{\upsilon + 1}{2}} \frac{1}{\sigma} dx, \quad u = \frac{x - \mu}{\sigma}, du = \frac{1}{\sigma}dx\\
&= \int^{u=\frac{1 - \mu}{\sigma}}_{u=\frac{0 - \mu}{\sigma}} \left(\sigma u + \mu \right)\frac{\Gamma\left(\frac{\upsilon +1}{2}\right)}{\sqrt{\upsilon \pi} \Gamma\left(\frac{\upsilon}{2}\right)}\left(1 + \frac{u^2}{\upsilon}\right)^{-\frac{\upsilon + 1}{2}} du\\
&= \int^{u=\frac{1 - \mu}{\sigma}}_{u=\frac{0 - \mu}{\sigma}} \sigma u f(u) du + \int^{u=\frac{1 - \mu}{\sigma}}_{u=\frac{0 - \mu}{\sigma}} \mu f(u) du\\
\end{split}
\end{equation}

Substituting this in Equation~\ref{eq:truncated_t_mean_base}, we have:
\begin{equation}
\begin{split}
&\frac{\int^{u=\frac{1 - \mu}{\sigma}}_{u=\frac{0 - \mu}{\sigma}} \sigma u f(u) du + \int^{u=\frac{1 - \mu}{\sigma}}_{u=\frac{0 - \mu}{\sigma}} \mu f(u) du}{F(\frac{1-\mu}{\sigma}) - F(\frac{0-\mu}{\sigma})}\\
&= \sigma \mathbb{E}\left[u\ |\ \frac{-\mu}{\sigma} \le u \le \frac{1-\mu}{\sigma}\right] + \mu
\end{split}
\end{equation}

\section{Bayesian analysis of BFTS}
\label{supplemental:bayesian_analysis}

This section performs a Bayesian analysis of BFTS \citep{Libin2019}, motivating its \textit{pure exploration} strategy.
We present two heuristics that form the basis of BFTS's exploration strategy, related to their probability of error.

In this Bayesian framework, we reason about the full distribution over bandits.
Consequently, the actual means $\boldsymbol{\mu}$ are unknown, and we assert our belief over $\boldsymbol{\mu}$ given 
\begin{equation}
  \label{eq:bandit_belief}
\pi(\cdot \ |\  \mathcal{H}^{(t-1)}), 
\end{equation}
i.e., the prior belief over the means $\pi(.)$ conditioned on the observed history 
\begin{equation}
  \label{eq:history}
\mathcal{H}^{(t-1)}= \left\{a^{(i)}, r^{(i)}\right\}^{(t-1)}_{i=1} 
\end{equation}
 at time $t$.

We define $\Psi_{\rank}(\boldsymbol{\theta}^{(t)})$ as the $\rank$ ordered arm.
We specify the random variables $A^{\pi}_{\rank}$ as the $\rank$-ranked arms according to the prior belief, and $A^{TS}_\rank$ as the $\rank$-ranked arm according to Thompson sampling (TS):
\begin{equation}
\begin{split}
&A^{\pi}_{\rank} = \Psi_{\rank}(\boldsymbol{\mu})\\
&A^{TS}_{\rank} = \Psi_{\rank}(\boldsymbol{\theta}^{(t)})
\end{split}
\end{equation}
As TS is a \emph{probability matching} algorithm \cite{agrawal2012analysis,russo2016information}, it samples directly from the belief asserted in Equation~\ref{eq:bandit_belief}. Formally, this is defined as:
\begin{equation}
P(A^{TS}_{\rank} = \cdot\ |\ \mathcal{H}^{(t-1)}) = P(A^{\pi}_{\rank} = \cdot\ |\ \mathcal{H}^{(t-1)})
\label{eq:prob_match}
\end{equation}

We define $\rank^+ \in [1, \dots, m]$ and $\rank^- \in [m+1, \dots, K]$.
Using this notation, we can express the true optimal arm set $J^*$ and recommended arm set $J^{TS}$ as:
\begin{equation}
\begin{split}
&J^* = \{A^{\pi}_{\rank^+} \ |\ \forall \rank^+\}  \\
&J^{TS} = \{ A^{TS}_{\rank^+} \ |\ \forall \rank^+\} ,
\end{split}
\label{eq:J_TS_and_J_star}
\end{equation}
we refer to $\overline{J^*}$ as the complement of $J^*$, i.e., the set of all arms excluding $J^*$.
Note that both $J^*$ and $J^{TS}$ are random variables, as they are expressed as a union of random variables. 
We use $P_t(.)$ to denote a probability that is conditioned on the observed history $\mathcal{H}^{(t-1)}$ at time $t$:
\begin{equation}
P_t(.) = P(. \ |\ \mathcal{H}^{(t-1)})
\end{equation}
Given this framework, we identify two heuristics that underlie BFTS' sampling strategy.

\begin{heuristic}
\label{heur:heur_1}
The expectation that BFTS wrongly ranks an arm that is believed to be optimal is bounded by the probability that BFTS wrongly ranks the arm on the sub-optimal side of the decision boundary:
\begin{equation}
\begin{split}
&\E_{\rank^-}[P_t(A^{TS}_{\rank^-} \in J^*)] \leq P_t(A^{TS}_{m+1} \in J^*)
\end{split}
\end{equation}
Given this inequality, we expect that sampling the $m\!+\!1$-th arm will reduce $\E_{\rank^-}[P_t(A^{TS}_{\rank^-} \in J^*)]$.
\end{heuristic}

\begin{heuristic}
\label{heur:heur_2}
The expectation that BFTS wrongly ranks an arm that is believed to be sub-optimal is bounded by the probability that BFTS wrongly ranks the arm ranked on the optimal side of the decision boundary.
\begin{equation}
\begin{split}
&\E_{\rank^+}[P_t(A^{TS}_{\rank^+} \in \overline{J^*})] \leq P_t(A^{TS}_{m} \in \overline{J^*})
\end{split}
\end{equation}
Given this inequality, we expect that sampling the $m$-th arm will reduce\\\noindent $\nolinebreak{\E_{\rank^+}[P_t(A^{TS}_{\rank^+} \in \overline{J^*})]}$. 
\end{heuristic}

These heuristics come from the fact that it is counterintuitive for TS to often order an arm as optimal when it is \emph{believed} to be suboptimal.
However, due to the stochastic nature of both the bandit and TS, it is possible to end up with a posterior for which the heuristics do not hold. 
Notwithstanding, we argue that given the intuition behind probability matching, such events become unlikely when reasonable priors are chosen and BFTS' exploration strategy is followed.

We now show how the expectations in the heuristics relate to the probability of error. 
As such, given the heuristics, we can bound the probability of error with respect to both sides of the decision boundary (i.e., $A^{TS}_{m+1}$ and $A^{TS}_m$), demonstrating that BFTS' exploration strategy is well-grounded.

First, we derive the bound in terms of $A^{TS}_{m+1}$:
\begin{equation}
\begin{split}
P_t\left(J^* \neq J^{TS}\right) &= P_t\left(\bigvee_{\rank^-}A^{TS}_{\rank^-} \in J^*\right)\\
	&\leq \sum_{\rank^-}P_t\left(A^{TS}_{\rank^-} \in J^*\right)\\
	&= \frac{\sum_{\rank^-}P_t\left(A^{TS}_{\rank^-} \in J^*\right) \cdot \left(K-m\right)}{\left(K-m\right)}\\
	&= \E_{\rank^-}\big[P_t\left(A^{TS}_{\rank^-} \in J^*\right)\big] \cdot \left(K-m\right)\\
	&\stackrel{(H\ref{heur:heur_1})}{\leq}  P_t\left(A^{TS}_{m+1} \in J^*\right) \cdot \left(K-m\right) 
\end{split}
\end{equation}
In the first step, we express the probability of error in terms of the arms that are ranked as sub-optimal by TS.
In the second step, we apply a union bound.
In the third and fourth step, we transform the sum to an expected value.
In the final step, we apply Heuristic~\ref{heur:heur_1} $(H\ref{heur:heur_1})$.

Following analogous arguments, we derive the bound in terms of $A^{TS}_{m}$ by applying Heuristic~\ref{heur:heur_2} (full derivation in Supplementary Information):
\begin{equation}
\begin{split}
&P_t\left(J^* \neq J^{TS}\right) \leq P_t\left(A^{TS}_{m} \in \overline{J^*}\right) \cdot m 
\end{split}
\end{equation}

These insights motivate a uniform selection of the two arms on both sides of the decision boundary, as is reflected in BFTS (see Algorithm~\ref{algorithm:BFTS}, lines 2 and 3 in the for loop).

The BFTS algorithm is constructed such that its sampling strategy is completely independent of its recommendation strategy.
Likewise, in this analysis, we consider the belief that BFTS maintains over the problem, in terms of the random variable $J^{TS}$ (Equation~\ref{eq:J_TS_and_J_star}), rather than the statistic that is used to make recommendations (e.g., the mean of the posterior in our experiments).
This observation shows that our analysis is independent from the statistic used to make recommendations with BFTS.

When inspecting other algorithms for the $m$-top setting, we observe that the decision boundary between the $m^\text{th}$ and ${m+1}^\text{th}$ arms also plays an important role. For example, the frequentist algorithm AT-LUCB samples two arms each step; the one with the smallest lower-bound among the $m$-top arms, and the one with the greatest upper-bound among the rest. This is analogous to choosing the optimal and sub-optimal arms that are closest to the decision boundary.

\section{STRIDE contact tracing configuration}
\label{supplemental:stride_configuration}

Based on configurations by Willem et al. \citep{Willem2021}, for contact tracing we assume a 0.7 detection probability with a daily case finding capacity of 10.000 individuals. Tests have a 0.1 probability to be false negative results. There is a 1 day delay for detecting symptomatic individuals and 2 days delay for isolating infected individuals.

\clearpage

\section{COVID-19 experiments}
\label{supplemental:stride_remaining_posteriors}

\subsection{65\% uptake proportion}

\begin{figure}[H]
    \centering
    \includegraphics[width=0.9\linewidth, keepaspectratio]{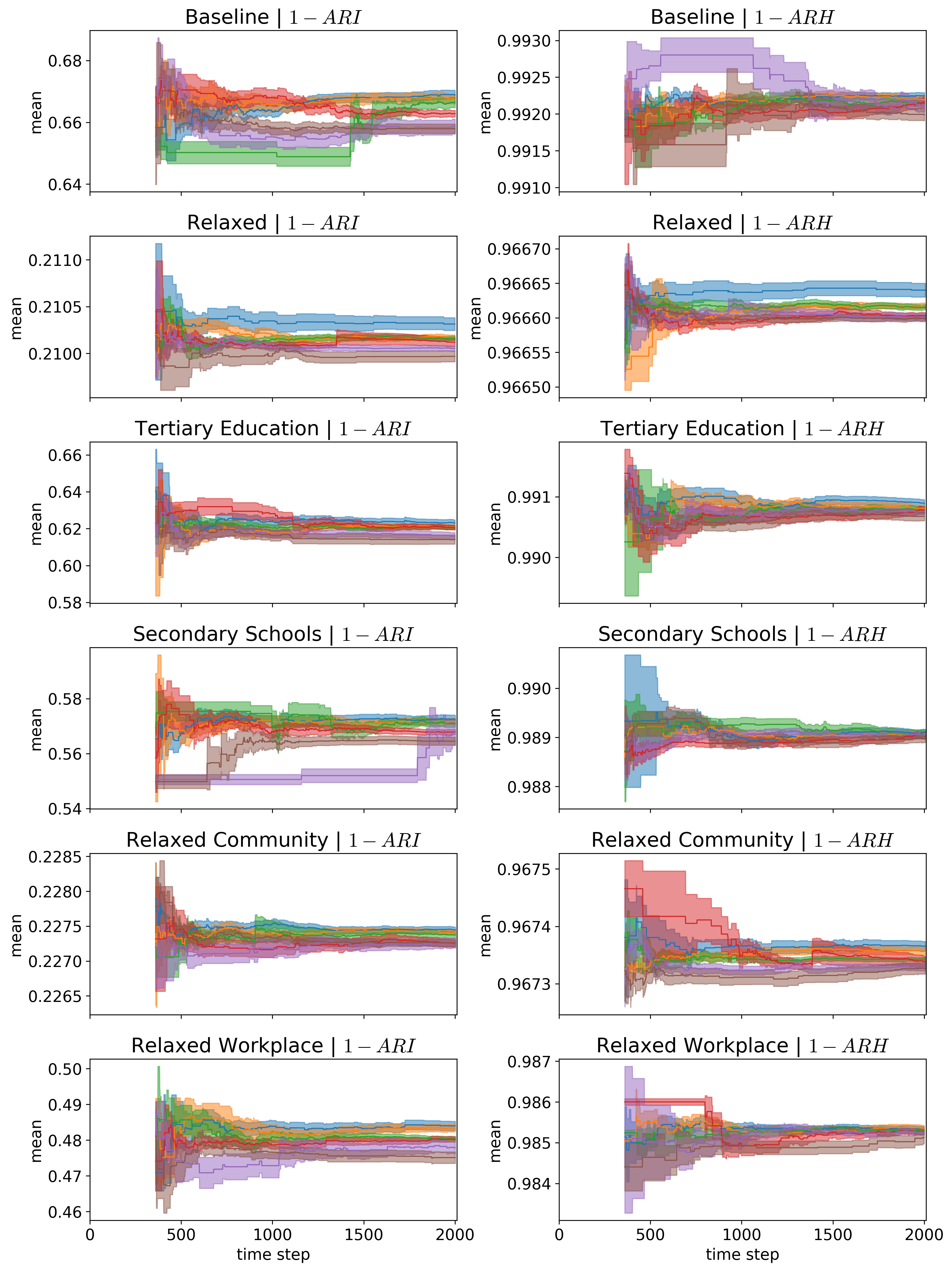}
    \caption{Estimated means and uncertainty (standard deviation) for the 6 arms around the decision boundary. (left) infections \infections. (right) hospitalisations \hospitalisations. For the top-$10$ strategies, these are the 8th-13th ranked arms under different contact reduction schemes (Table \ref{table:configs}), for a 65\% vaccine uptake proportion.}
    \label{figure:stride_posteriors_b0.65}
\end{figure}

\subsection{70\% uptake proportion}

\begin{figure}[H]
    \centering
    \includegraphics[width=0.9\linewidth, keepaspectratio]{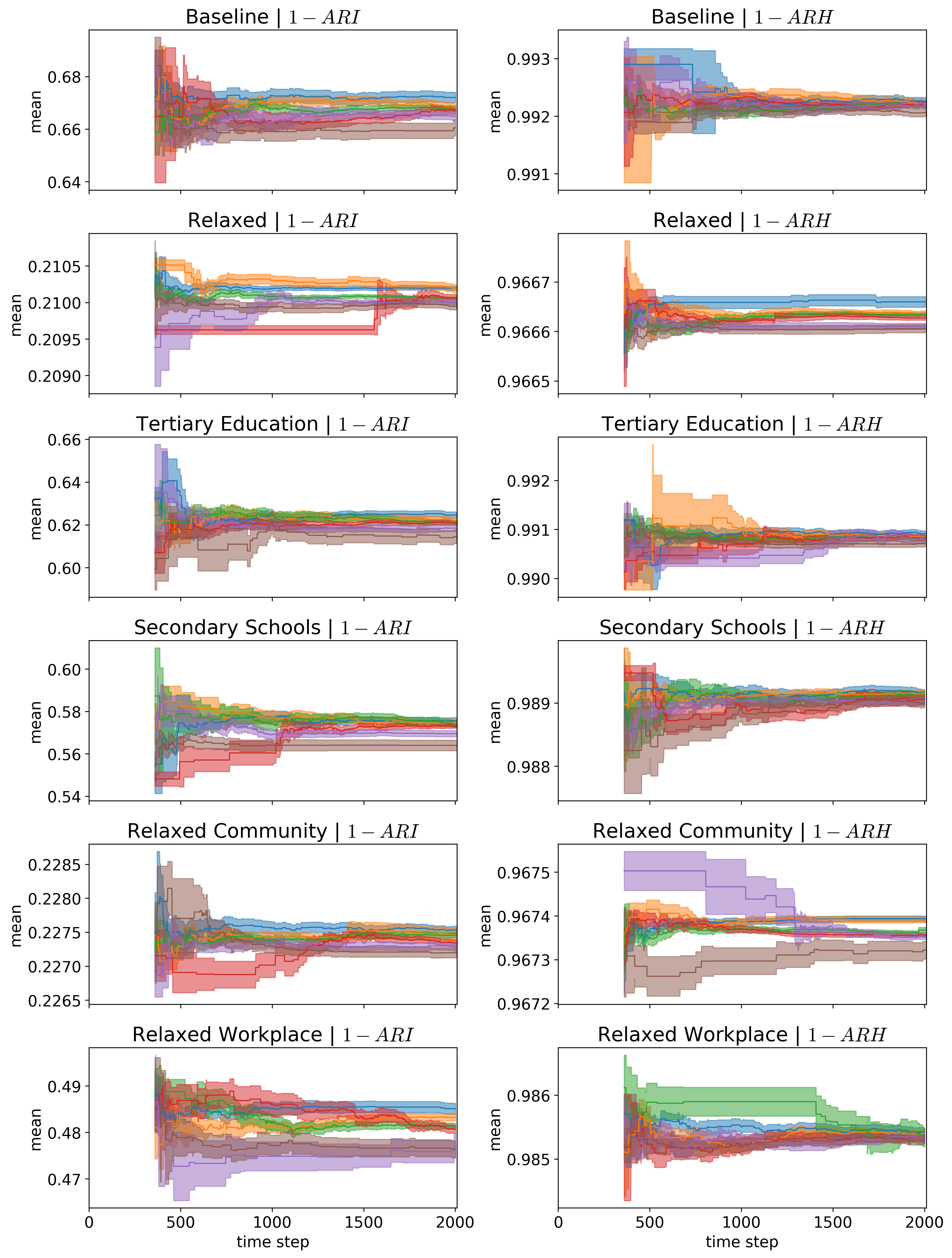}
    \caption{Estimated means and uncertainty (standard deviation) for the 6 arms around the decision boundary. (left) \infections. (right) \hospitalisations. For the top-$10$ top strategies, these are the 8th-13th ranked arms under different contact reduction schemes (Table \ref{table:configs}), for a 70\% vaccine uptake proportion.}
    \label{figure:stride_posteriors_b0.7}
\end{figure}

\subsection{75\% uptake proportion}

\begin{figure}[H]
    \centering
    \includegraphics[width=0.9\linewidth, keepaspectratio]{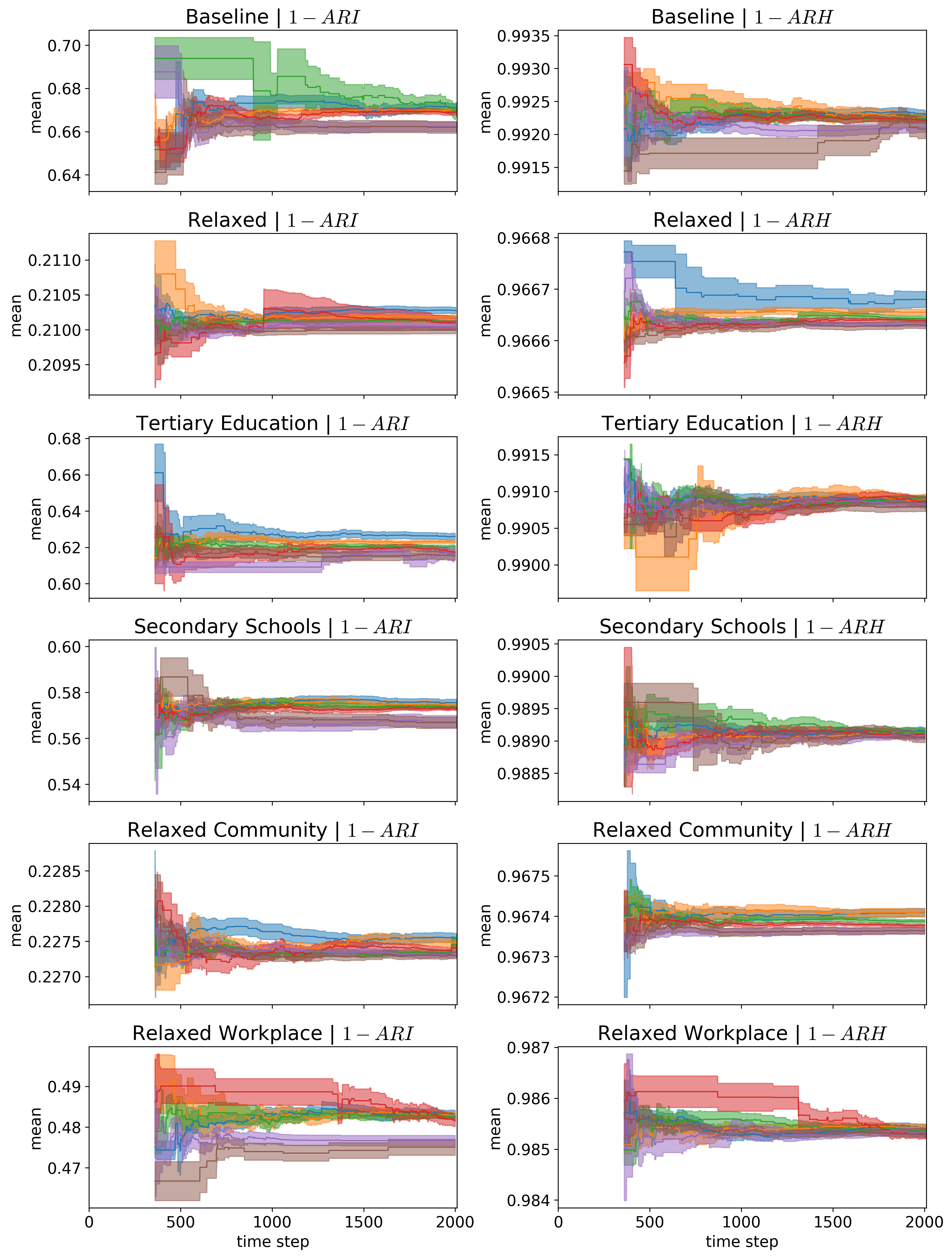}
    \caption{Estimated means and uncertainty (standard deviation) for the 6 arms around the decision boundary. (left) \infections. (right) \hospitalisations. For the top-$10$ strategies, these are the 8th-13th ranked arms under different contact reduction schemes (Table \ref{table:configs}), for a 75\% vaccine uptake proportion.}
    \label{figure:stride_posteriors_b0.75}
\end{figure}

\subsection{80\% uptake proportion}

\begin{figure}[H]
    \centering
    \includegraphics[width=0.9\linewidth, keepaspectratio]{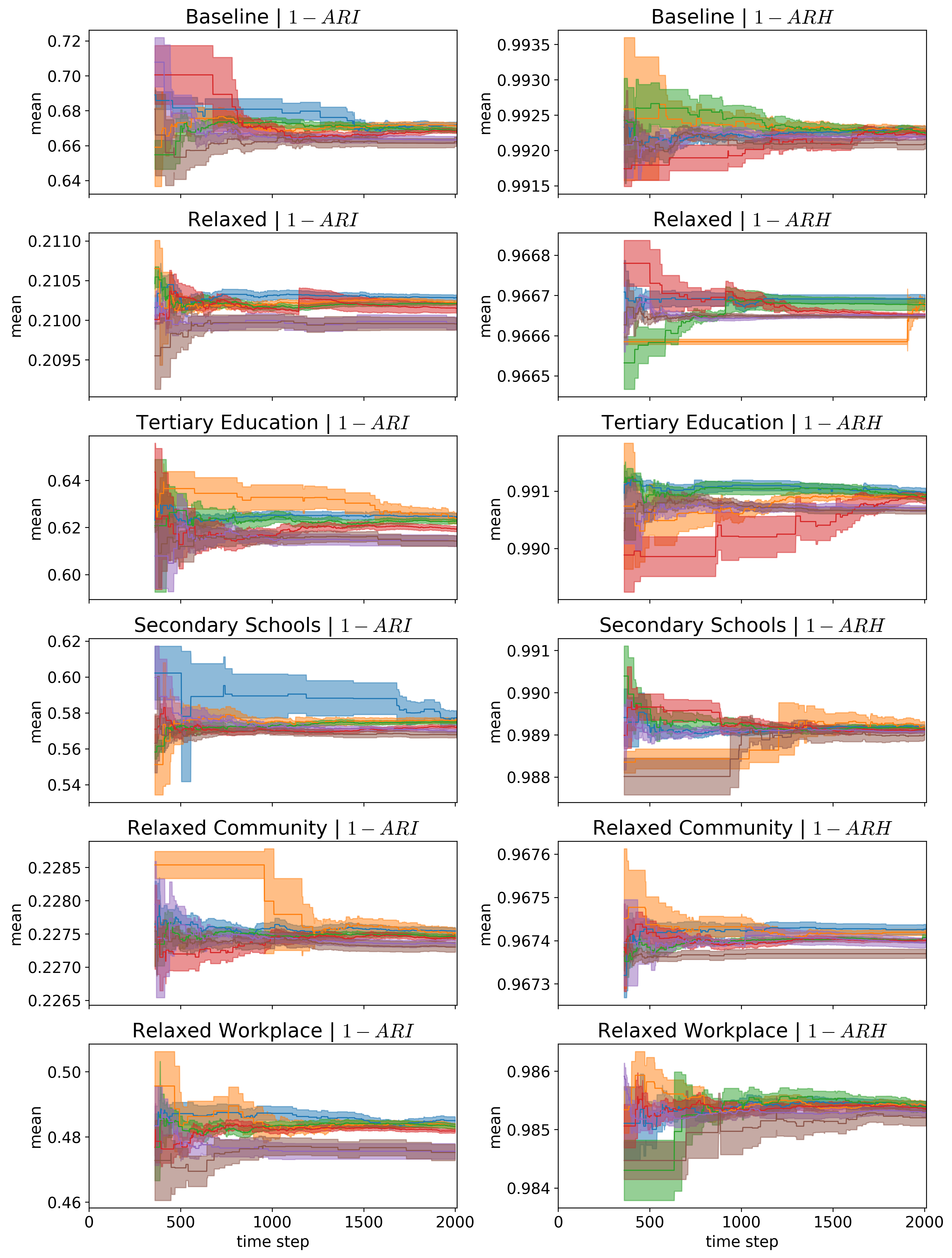}
    \caption{Estimated means and uncertainty (standard deviation) for the 6 arms around the decision boundary. (left) \infections. (right) \hospitalisations. For the top-$10$  strategies, these are the 8th-13th ranked arms under different contact reduction schemes (Table \ref{table:configs}), for a 80\% vaccine uptake proportion.}
    \label{figure:stride_posteriors_b0.8}
\end{figure}

\subsection{85\% uptake proportion}

\begin{figure}[H]
    \centering
    \includegraphics[width=0.9\linewidth, keepaspectratio]{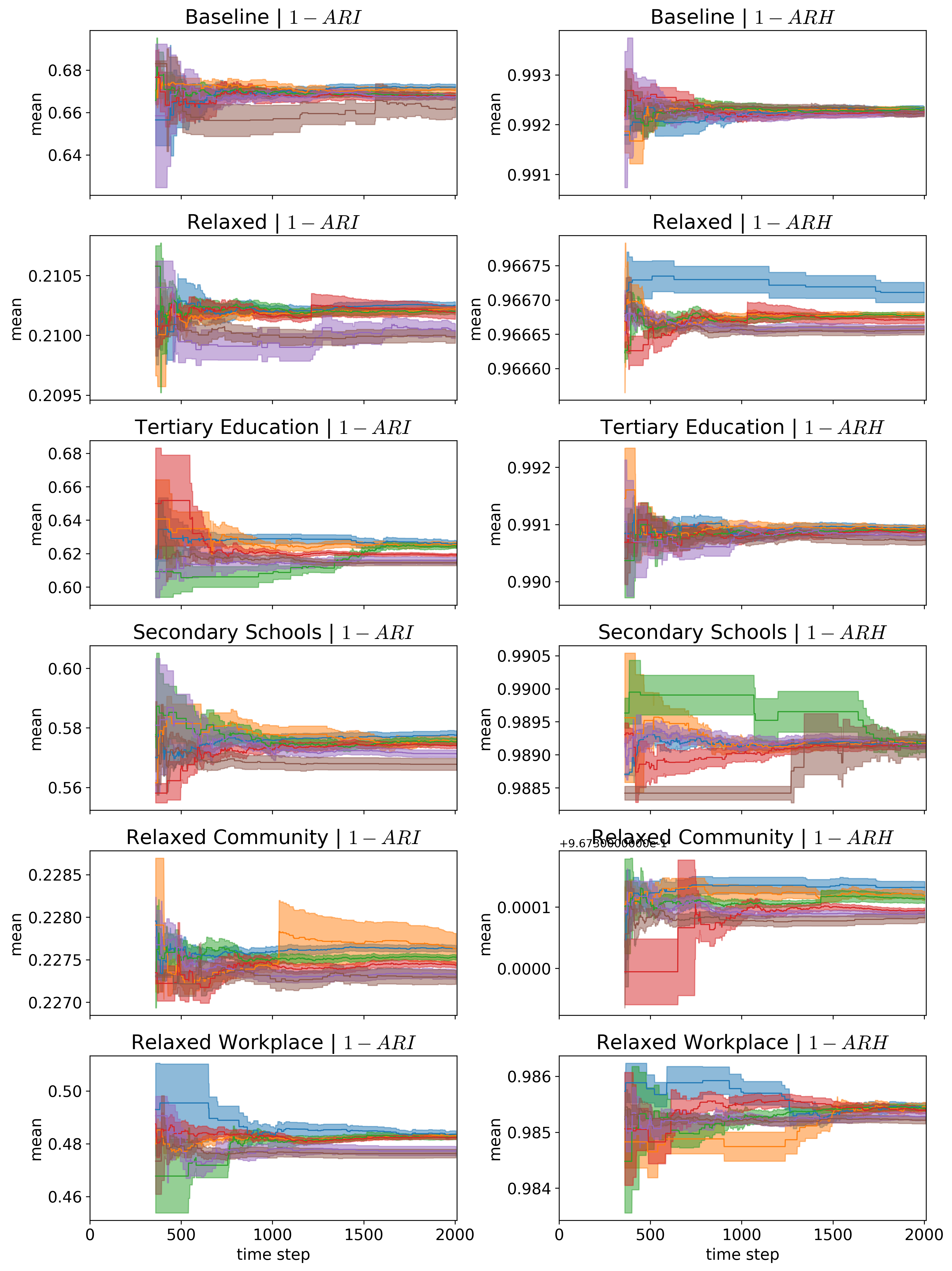}
    \caption{Estimated means and uncertainty (standard deviation) for the 6 arms around the decision boundary. (left) \infections. (right) \hospitalisations. For the top-$10$ strategies, these are the 8th-13th ranked arms under different contact reduction schemes (Table \ref{table:configs}), for a 85\% vaccine uptake proportion.}
    \label{figure:stride_posteriors_b0.85}
\end{figure}

\section{Top-$10$ vaccination strategies under different uptake percentages}
\label{supplemental:stride_remaining_top_uptakes}

\subsection{65\% uptake proportion}

\begin{figure}[H]
    \centering
    \includegraphics[width=1.0\linewidth, keepaspectratio]{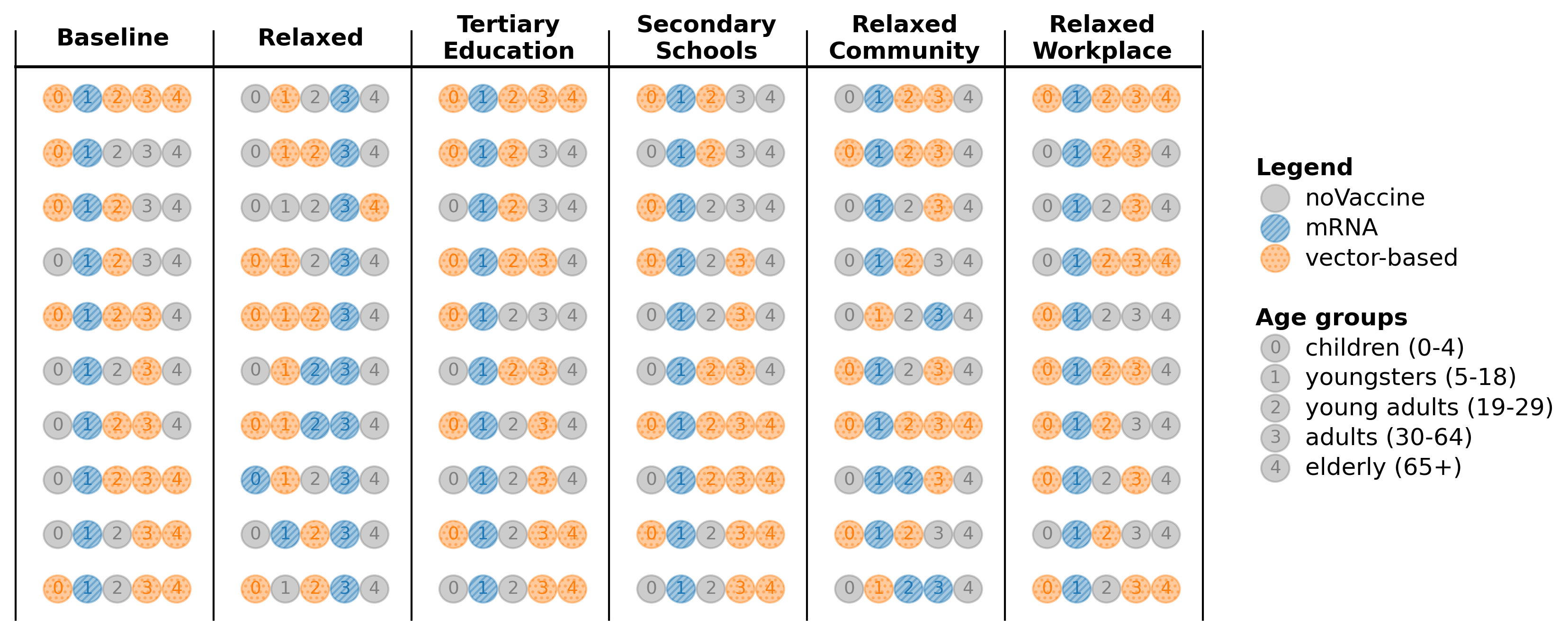}
    \caption{Learned top-$10$ vaccination strategies when minimising the infection attack rate (\infections) under various contact reduction schemes, under a 65\% vaccine uptake proportion.}
    \label{figure:config_rankings_inf0.65}
\end{figure}

\begin{figure}[H]
    \centering
    \includegraphics[width=1.0\linewidth, keepaspectratio]{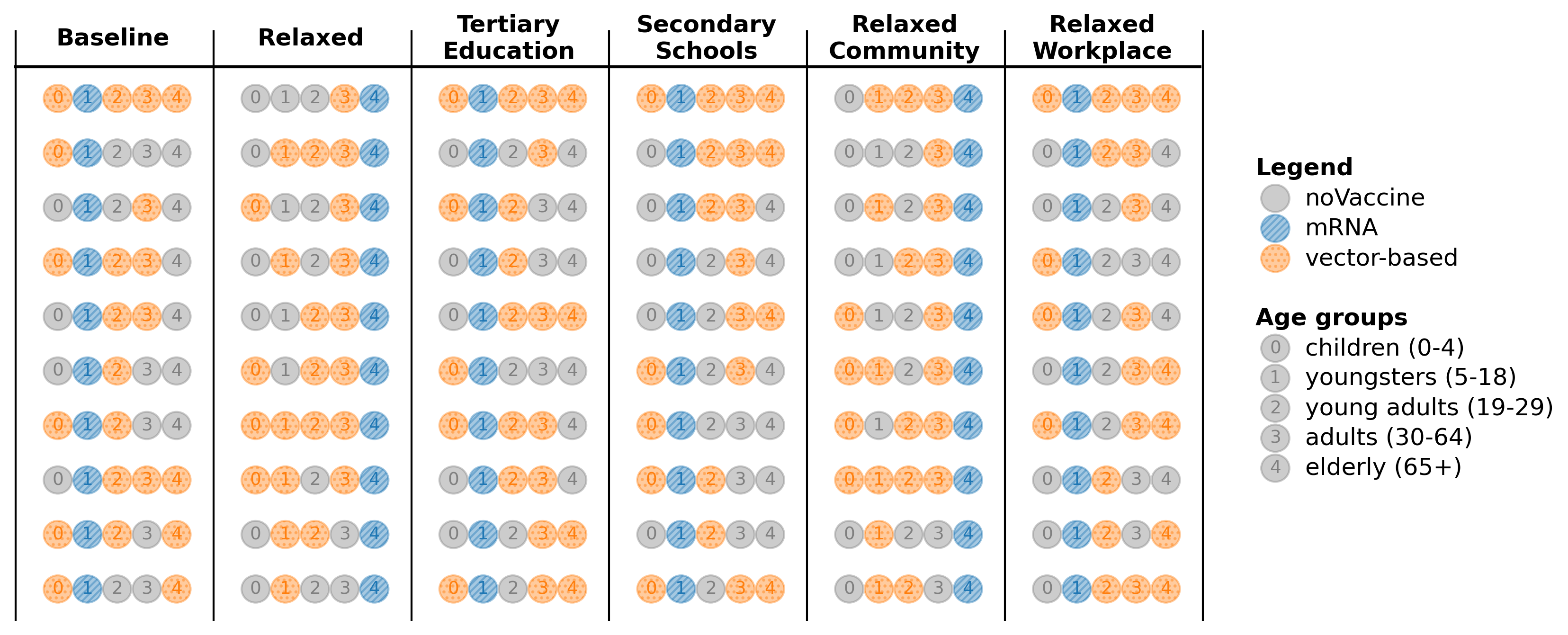}
    \caption{Learned top-$10$ vaccination strategies when minimising the hospitalisation attack rate (\hospitalisations) under various contact reduction schemes, under a 65\% vaccine uptake proportion.}
    \label{figure:config_rankings_hosp0.65}
\end{figure}

\subsection{70\% uptake proportion}

\begin{figure}[H]
    \centering
    \includegraphics[width=1.0\linewidth, keepaspectratio]{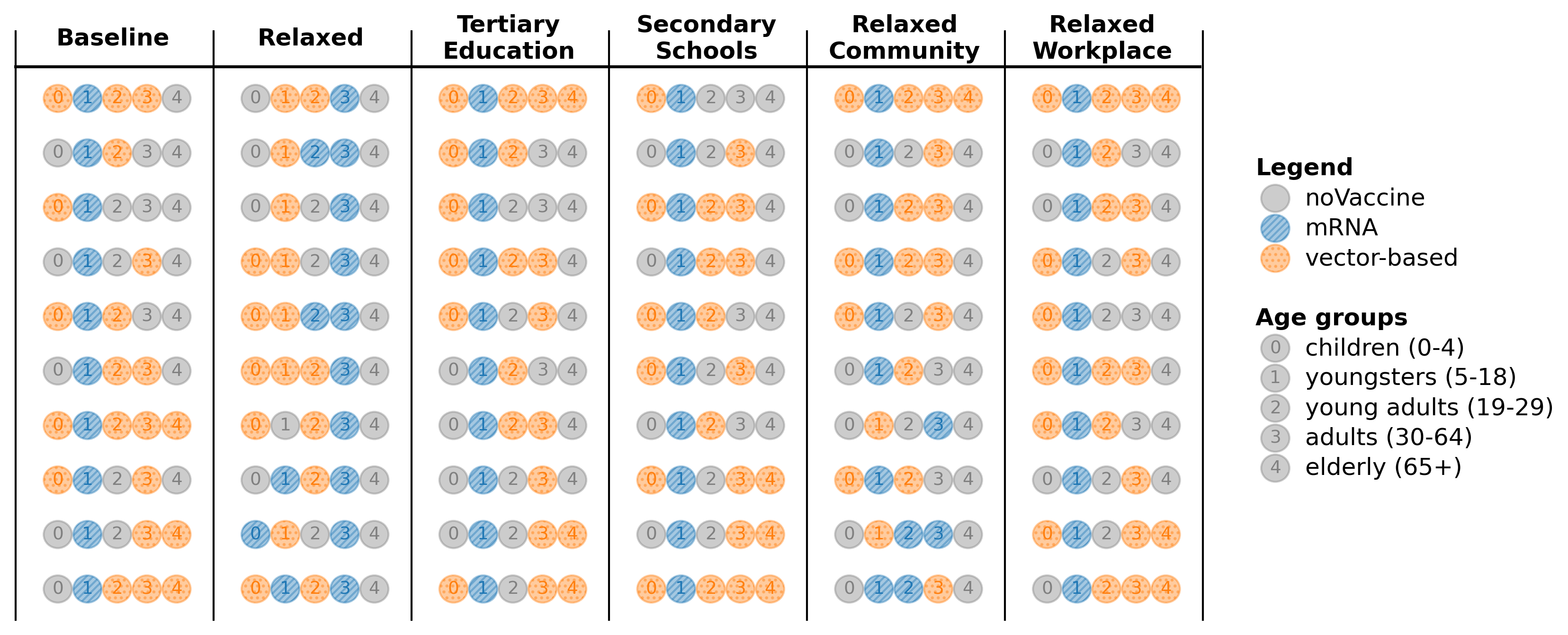}
    \caption{Learned top-$10$ vaccination strategies when minimising the infection attack rate (\infections) under various contact reduction schemes, under a 70\% vaccine uptake proportion.}
    \label{figure:config_rankings_inf0.7}
\end{figure}

\begin{figure}[H]
    \centering
    \includegraphics[width=1.0\linewidth, keepaspectratio]{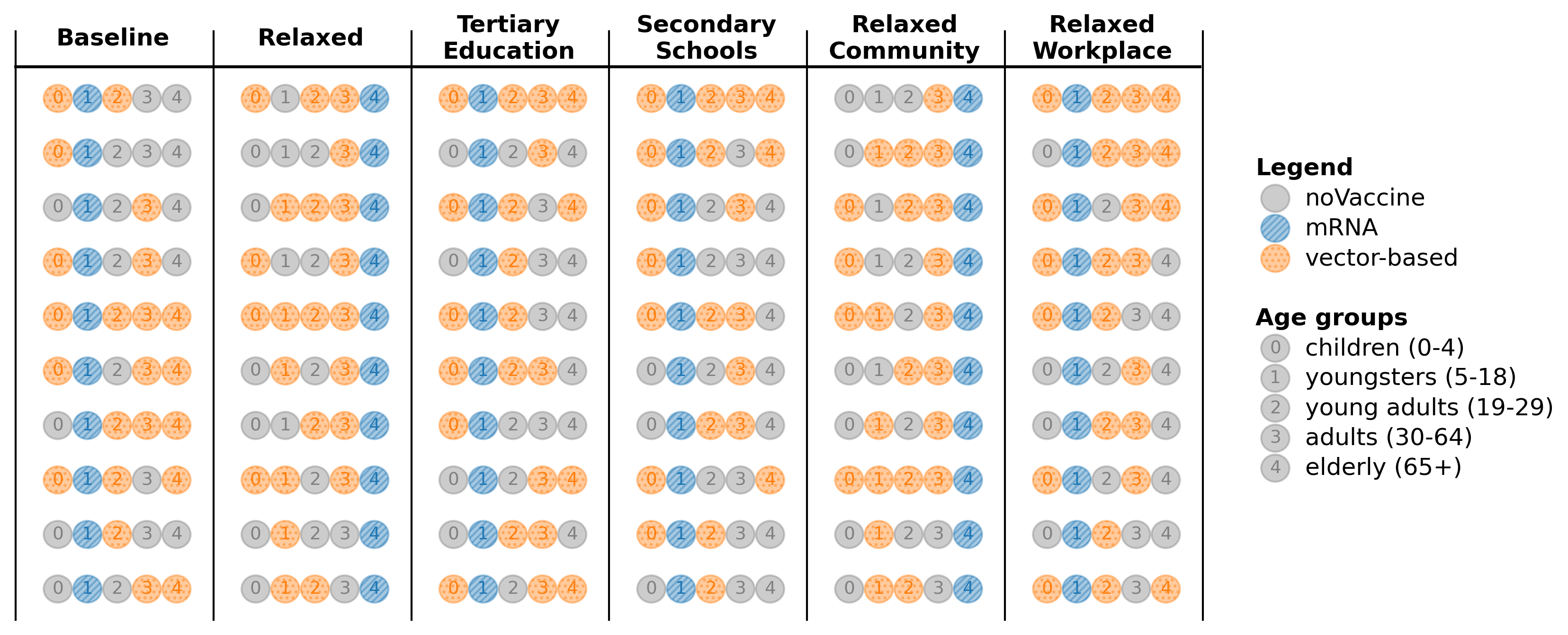}
    \caption{Learned top-10 vaccination strategies when minimising the hospitalisation attack rate (\hospitalisations) under various contact reduction schemes, under a 70\% vaccine uptake proportion.}
    \label{figure:config_rankings_hosp0.7}
\end{figure}

\subsection{80\% uptake proportion}

\begin{figure}[H]
    \centering
    \includegraphics[width=1.0\linewidth, keepaspectratio]{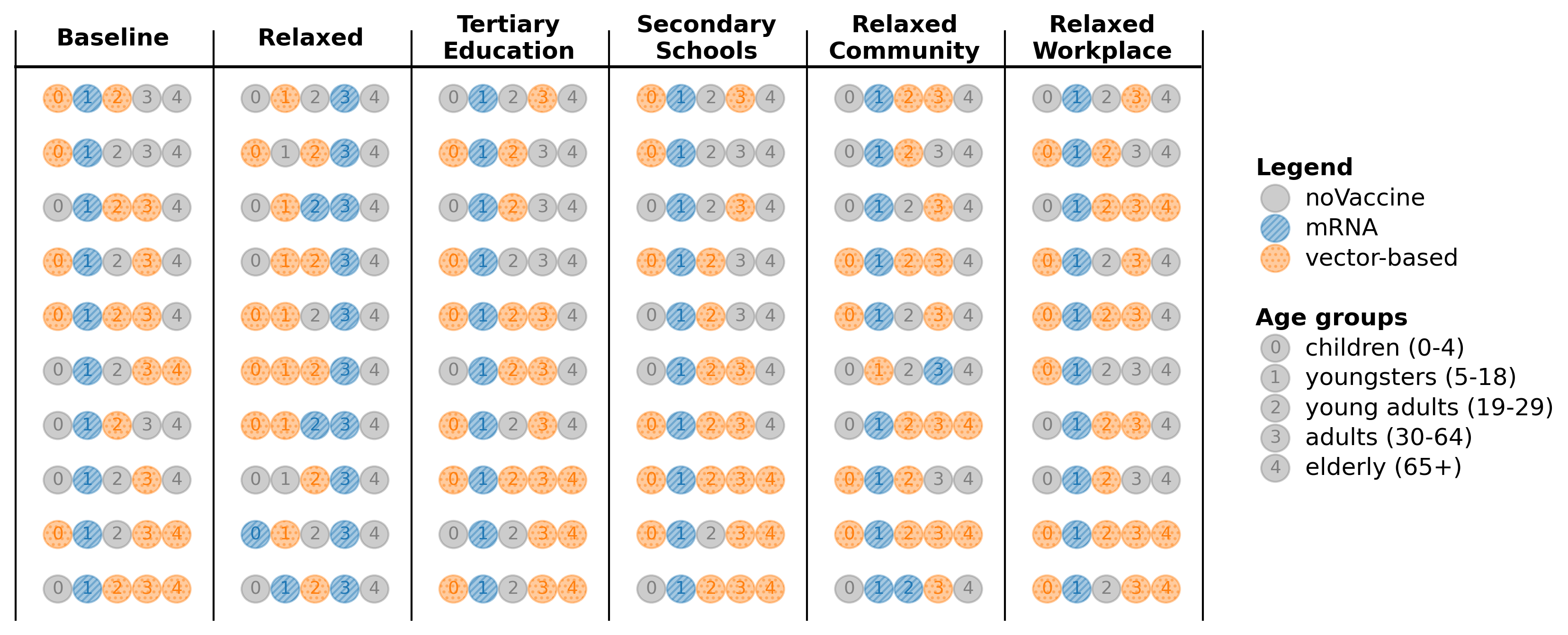}
    \caption{Learned top-10 vaccination strategies when minimising the infection attack rate (\infections) under various contact reduction schemes, under a 80\% vaccine uptake proportion.}
    \label{figure:config_rankings_inf0.8}
\end{figure}

\begin{figure}[H]
    \centering
    \includegraphics[width=1.0\linewidth, keepaspectratio]{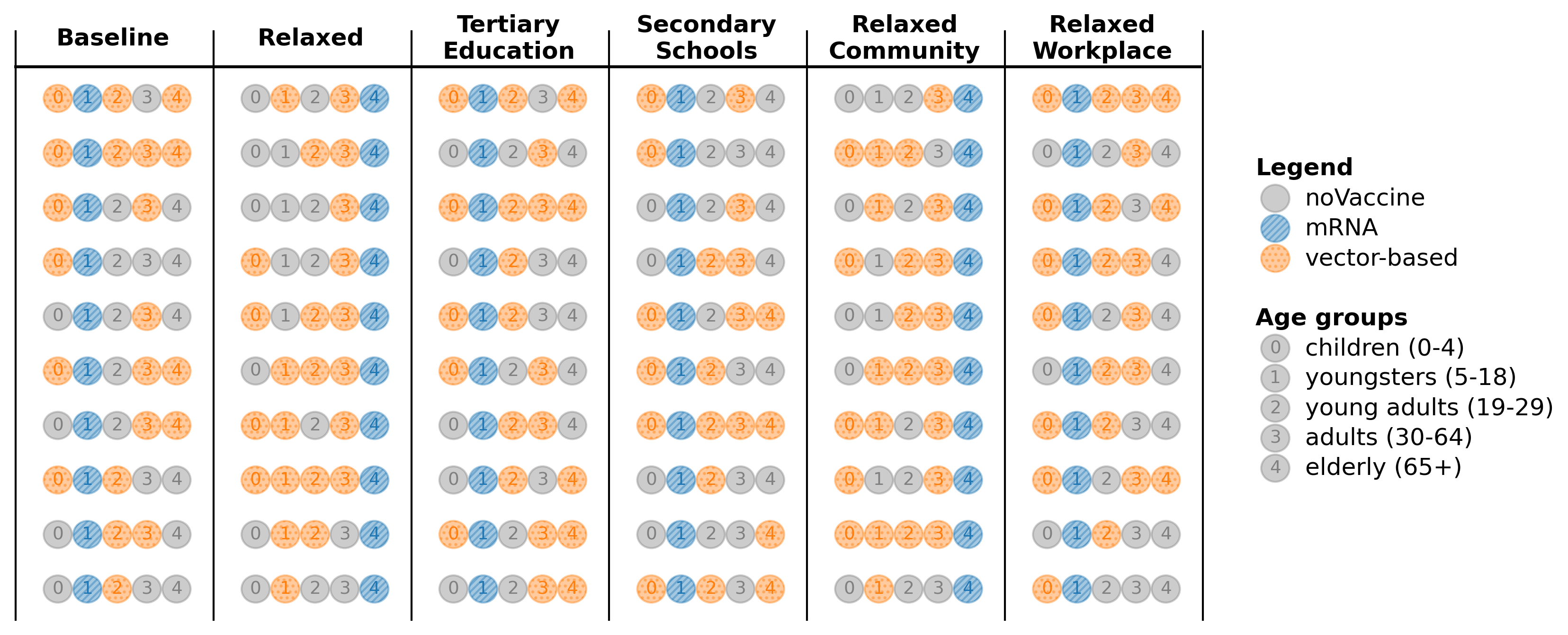}
    \caption{Learned top-10 vaccination strategies when minimising the hospitalisation attack rate (\hospitalisations) under various contact reduction schemes, under a 80\% vaccine uptake proportion.}
    \label{figure:config_rankings_hosp0.8}
\end{figure}

\subsection{85\% uptake proportion}

\begin{figure}[H]
    \centering
    \includegraphics[width=1.0\linewidth, keepaspectratio]{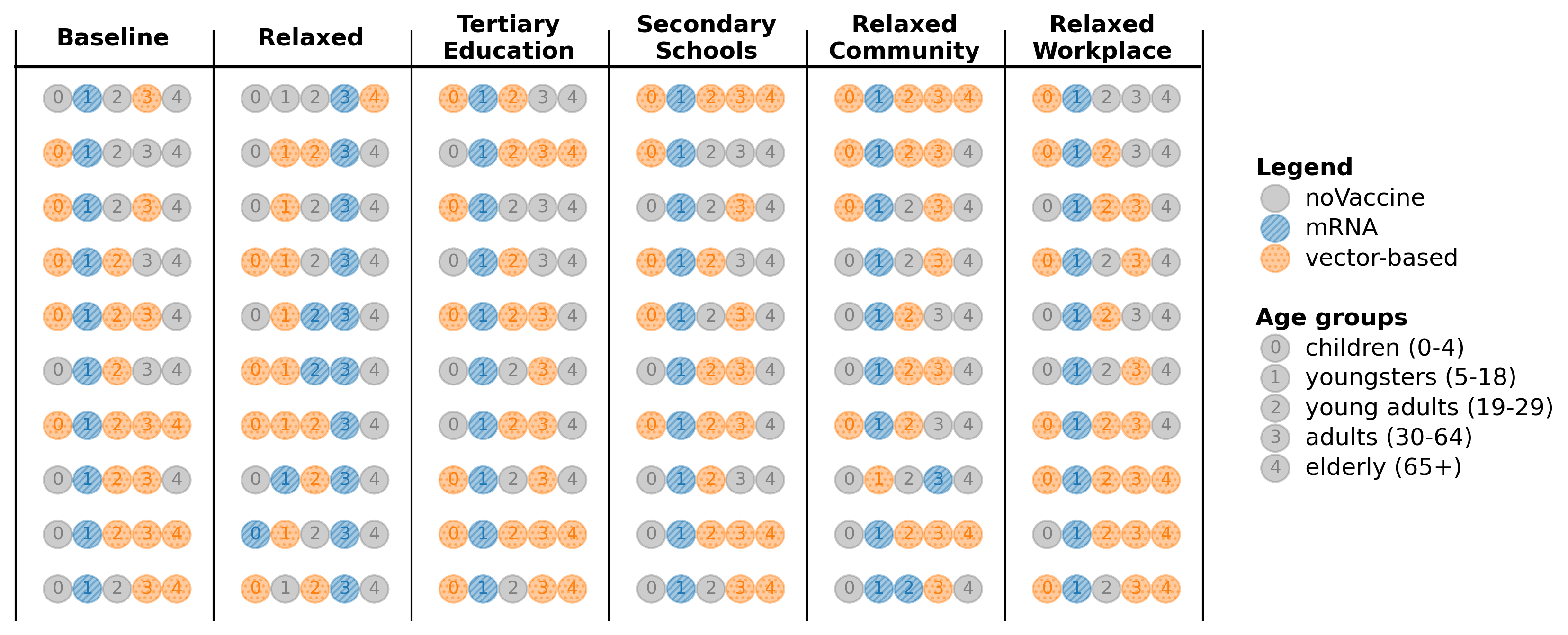}
    \caption{Learned top-10 vaccination strategies when minimising the infection attack rate (\infections) under various contact reduction schemes, under a 85\% vaccine uptake proportion.}
    \label{figure:config_rankings_inf0.85}
\end{figure}

\begin{figure}[H]
    \centering
    \includegraphics[width=1.0\linewidth, keepaspectratio]{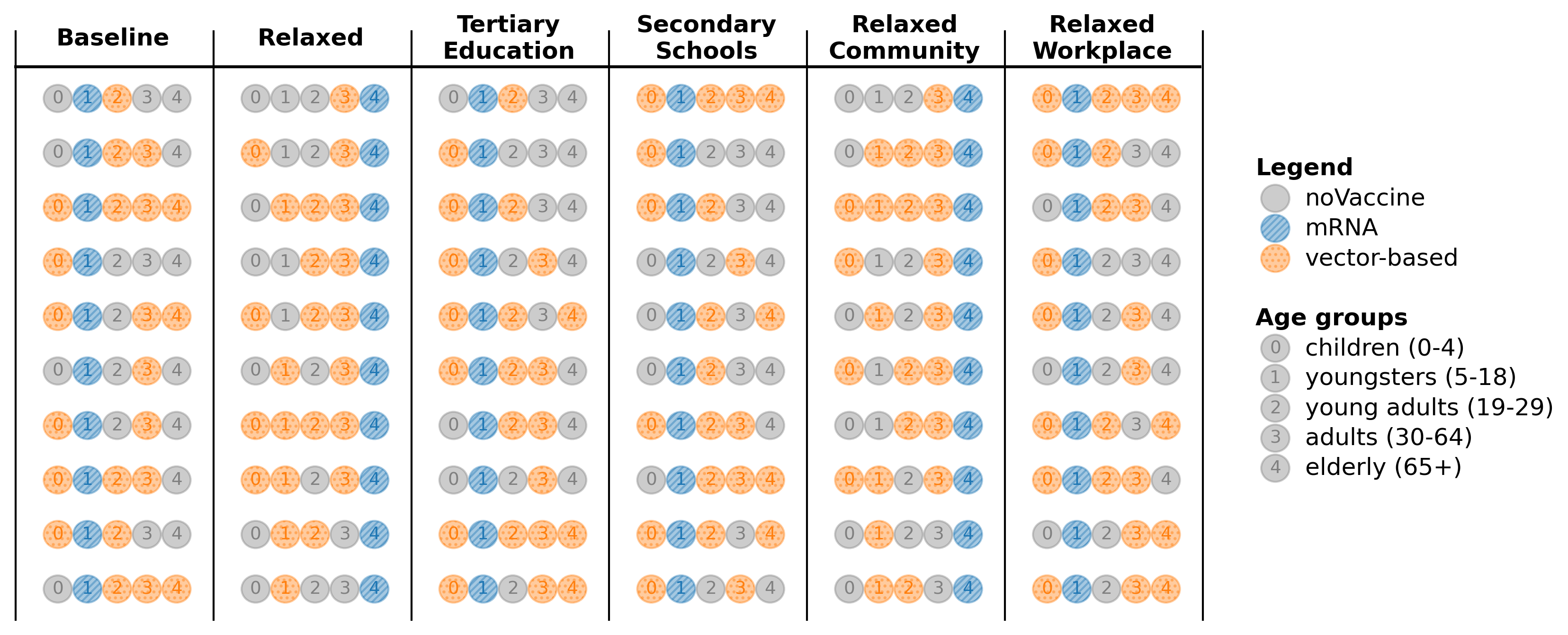}
    \caption{Learned top-$10$ vaccination strategies when minimising the hospitalisation attack rate (\hospitalisations) under various contact reduction schemes, under a 85\% vaccine uptake proportion.}
    \label{figure:config_rankings_hosp0.85}
\end{figure}

\end{appendices}

\end{document}